\documentclass[10pt,twocolumn,letterpaper]{article}

\usepackage{iccv}
\usepackage{times}
\usepackage{epsfig}
\usepackage{graphicx}
\usepackage{booktabs}
\usepackage{amsmath, amsthm, amssymb}

\usepackage{algorithm}
\usepackage{algorithmic}

\usepackage{graphicx}
\usepackage{bbm}
\usepackage{bbding}
\usepackage{url}
\usepackage{threeparttable}

\usepackage{caption}
\usepackage{balance}
\usepackage{subcaption}
\usepackage{colortbl}
\usepackage{multirow}

\usepackage{color}

\def\ie{\textit{i.e.}}
\def\eg{\textit{e.g.}}
\usepackage[pagebackref,breaklinks,colorlinks]{hyperref}

\usepackage{lipsum}

\newcommand\blfootnote[1]{%
  \begingroup
  \renewcommand\thefootnote{}\footnote{#1}%
  \addtocounter{footnote}{-1}%
  \endgroup
}
\usepackage{authblk}
\usepackage[capitalize]{cleveref}
\crefname{section}{Sec.}{Secs.}
\Crefname{section}{Section}{Sections}
\Crefname{table}{Table}{Tables}
\crefname{table}{Tab.}{Tabs.}

\iccvfinalcopy 

\def\iccvPaperID{****} 

\ificcvfinal\fi

\begin{document}

\title{Downstream-agnostic Adversarial Examples}

\author{Ziqi Zhou\textsuperscript{\rm $*$}\textsuperscript{1 2 3 4}, Shengshan Hu\textsuperscript{\rm $*$}\textsuperscript{1 2 3 4},  Ruizhi Zhao\textsuperscript{\rm $*$}\textsuperscript{1 2 3 4}\\ \vspace{-0.5em} Qian Wang\textsuperscript{\rm $\ddagger$}, Leo Yu Zhang\textsuperscript{\rm $\mathsection$},   Junhui Hou\textsuperscript{\rm $\mathparagraph$}, Hai Jin\textsuperscript{\rm $\dagger$}\textsuperscript{1 2 5}\\\vspace{0.5em}

\textsuperscript{\rm $*$}School of Cyber Science and Engineering, Huazhong University of Science and Technology\\
\textsuperscript{\rm $\dagger$}School of Computer Science and Technology, Huazhong University of Science and Technology\\
\textsuperscript{\rm $\ddagger$}School of Cyber Science and Engineering, Wuhan University\\
\textsuperscript{\rm $\mathsection$}School of Information and Communication Technology, Griffith University\quad\\
\textsuperscript{\rm $\mathparagraph$}Department of Computer Science, City University of Hong Kong \\
{\tt\small \{zhouziqi, hushengshan, zhaoruizhi, hjin\}@hust.edu.cn}\\ 
{\tt\small qianwang@whu.edu.cn}, 
{\tt\small leo.zhang@griffith.edu.au}, 
{\tt\small jh.hou@cityu.edu.hk}
}

\maketitle
\ificcvfinal\thispagestyle{empty}\fi

\begin{abstract}
Self-supervised learning usually uses a large amount of unlabeled data to pre-train an encoder which can be used as a general-purpose feature extractor, such that downstream  users only need to perform fine-tuning operations to enjoy the benefit of ``large model". Despite this promising prospect, the security of pre-trained encoder has not been thoroughly investigated yet, especially when the pre-trained encoder is publicly available for commercial use. 
\vspace{0.2em}
\blfootnote{
\textsuperscript{1}National Engineering Research Center for Big Data Technology and System
\textsuperscript{2}Services Computing Technology and System Lab \
\textsuperscript{3}Hubei Key Laboratory of Distributed System Security \ 
\textsuperscript{4}Hubei Engineering Research Center on Big Data Security \ \textsuperscript{5}Cluster and Grid Computing Lab
}

In this paper, we propose AdvEncoder, the first framework for generating downstream-agnostic universal adversarial examples based on the pre-trained encoder. 
AdvEncoder aims to construct a universal adversarial perturbation or patch for a set of natural images that can fool all the downstream tasks inheriting the victim pre-trained encoder. 
Unlike traditional adversarial example works, the pre-trained encoder only outputs feature vectors rather than classification labels. Therefore, we first exploit the high frequency component information of the image to guide the generation of adversarial examples. 
Then we design a generative attack framework to construct adversarial perturbations/patches by learning the distribution of the attack surrogate dataset to improve their attack success rates and transferability.
Our results show that an attacker can successfully attack downstream tasks without knowing either the pre-training dataset or the downstream dataset.
We also tailor four defenses for pre-trained encoders, the results of which further prove the attack ability of AdvEncoder.
Our codes are available at: \url{https://github.com/CGCL-codes/AdvEncoder}.

\end{abstract}
\vspace{-0.6em}

\section{Introduction}
 \begin{figure}[!t]
  \setlength{\belowcaptionskip}{-0.5cm}  
    \centering
    \includegraphics[scale=0.58]{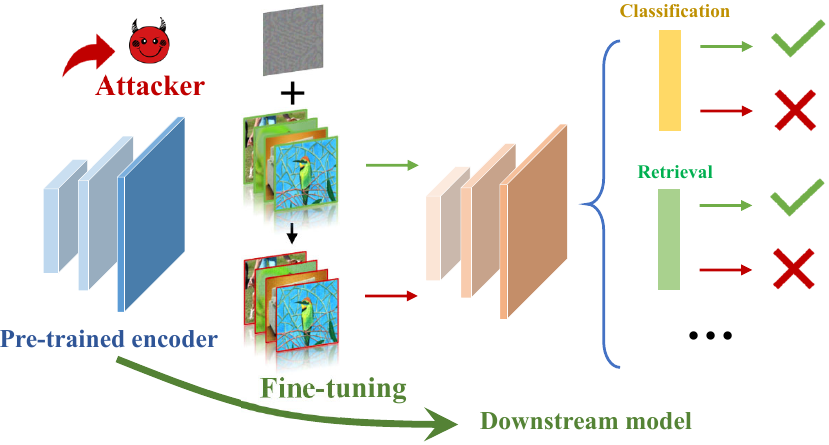}
    \caption{An overview of adversarial examples against different downstream tasks based on a pre-trained encoder
    }
    \label{fig:demo}
\end{figure}

\emph{Self-supervised learning}~\cite{chen2020simple, chen2020improved} (SSL) is an emerging machine learning paradigm that seeks to overcome the restrictions of labeled data. It usually uses a large volume of unlabeled data to pre-train a general-purpose encoder, which can be used as a feature extractor for various downstream tasks like image classification, image retrieval, object detection, etc. 
As a result, any resource-constrained user can enjoy the advantages of ``large model" without performing the expensive training from scratch, where only light-weight fine-tuning operations are needed at its request. 
Driven by this promising prospect, pre-training encoders become popular in industry and many service providers
publicly release their pre-trained encoders  (\eg, SimCLR by Google~\cite{chen2020simple, chen2020big}, MoCo by Meta~\cite{chen2021empirical, he2020momentum}) or deploy them as a commercial service (\eg, OpenAI~\cite{openaiapi},  Clarifai~\cite{clarifaiapi}).


Meanwhile, it is well known that \emph{deep neural networks} (DNNs) are vulnerable to various adversarial attacks~\cite{goodfellow2014explaining, moosavi2017universal, yang2020design, zhou2023advclip}, which will make pre-trained encoder fragile as well. However, the security of pre-trained encoder has received much less consideration in the literature.  
Although some recent works studied security threats on pre-trained encoders including backdoor attack~\cite{hu2022badhash, jia2022badencoder}, poisoning attack~\cite{liu2022poisonedencoder}, and privacy risks~\cite{cong2022sslguard, liu2021encodermi}, 
none of them 
paid attention to adversarial examples, another kind of prevalent and destructive attack on DNNs. 
Constructing adversarial examples against pre-trained encoders is quite different  from its traditional attack route due to the fact that the attacker has no knowledge of the downstream tasks. 
In other words, the attacker needs to attack a DNN  without knowing its task type, the pre-training dataset, and the downstream dataset,  even when the whole model will get fine-tuned.
\textit{To the best of our knowledge, how to realize adversarial example attack in the practical scenario of pre-training still remains challenging and unresolved.}

In this work, we take a big step towards bridging the gap between adversarial examples and pre-trained encoders. 
We consider both adversarial perturbation~\cite{carlini2017towards, goodfellow2014explaining, kurakin2018adversarial, moosavi2016deepfool}  and  patch~\cite{brown2017adversarial,hu2021advhash, liu2020bias, yang2020design}. The former one has a high imperceptibility, while the latter one is visible but confined to a small area of the image and more readily applicable in the physical world.
Furthermore, without the knowledge of downstream data,
we aim to realize universal adversarial attacks~\cite{hayes2018learning,moosavi2017universal,zhang2024whydoes} where  one adversarial perturbation or patch applies to a set of natural images and can cause model misclassification.

Specifically, we propose AdvEncoder, a novel attack framework for generating downstream-agnostic  universal adversarial examples.
The most challenging job lies in addressing the limitations and lacking supervised signals and the information about the downstream tasks.
Inspired by the fact that deep neural networks are biased towards texture features of images~\cite{jo2017measuring, wang2020high}, the change of texture information, 
\ie, the \emph{high  frequency components} (HFC) of the image,  is very likely to cause the model decision change. 
We first exploit a high frequency component filter to get the HFC of benign and adversarial samples, and pull away their Euclidean distance as much as possible to influence the model's decision. 
We then design a  generative attack framework to construct adversarial perturbations or patches with high attack success rates and transferability by learning the distribution of the data, with a fixed random noise as input. 
Our main contributions are summarized as follows:

\begin{itemize}
\item We propose AdvEncoder, the first attack framework to construct downstream-agnostic universal adversarial examples in self-supervised learning. We reveal that the pre-trained encoder incurs severe security risks for the downstream tasks.
\item We design a frequency-based generative network to generate universal adversarial examples by directly alearting  the texture  features of the image itself. It is a flexible framework that  can generate both adversarial perturbations and patches.
\item Our extensive experiments on fourteen  self-supervised training methods and four image datasets show that our AdvEncoder achieves high attack success rates and transferability against different downstream tasks. 
\item We tailor four popular defenses to mitigate AdvEncoder. The
results further prove the attack ability of AdvEncoder and highlight the needs of new defense mechanism to defend pre-trained encoders. 

\end{itemize}

\section{Background and Related Work}
\subsection{Self-supervised Learning}
Self-supervised learning seeks to utilize the oversight signals within the unlabeled data itself to pre-train encoders that can convert complex inputs into generic representations. The pre-trained encoder that learned generally valuable domain knowledge can be used as a universal feature extractor to transfer knowledge to solve different specific downstream tasks. In this paper, we concentrate on image encoders.

Based on~\cite{fini2022self, tao2022exploring},  self-supervised learning schemes can be divided into the following categories: \emph{(1) contrastive learning methods} (\eg, MoCo~\cite{chen2020improved, chen2021empirical}, SimCLR~\cite{chen2020simple}) train representations such that dissimilar negative pairs are widely apart and comparable positive pairs are shown to be near to one another. \emph{(2) negative-free methods} (\eg, BYOL~\cite{grill2020bootstrap}, Sim-Siam~\cite{chen2021exploring}, and ReSSL~\cite{zheng2021ressl}) achieve better representation without the use of negative samples by maintaining the consistency between positive samples and ignoring negative ones. \emph{(3) clustering-based methods} (\eg, SwAV~\cite{caron2020unsupervised}, DeepCluster v2~\cite{caron2020unsupervised}, and DINO~\cite{caron2021emerging}) group  similar samples into the same class using conventional clustering methods. \emph{(4) redundancy reduction-based methods} (\eg, Barlow Twins~\cite{zbontar2021barlow}, W-MSE~\cite{ermolov2021whitening}, VICReg~\cite{bardes2021vicreg}, and VIbCReg~\cite{lee2021vibcreg}) enhance the connection in the same dimension of the representation while attempting to decoupling in distinct dimensions. Concurrently, the use of nearest-neighbor retrieval has been investigated in NNCLR~\cite{dwibedi2021little}. These approaches start from different motivations, design different loss functions, and use different network structures and tricks, which also make them have different defense abilities against adversarial attacks.

\subsection{Attacks on Pre-trained Encoders}

Recently, a growing number of works began to investigate the privacy and security issues of the pre-trained encoders in self-supervised learning. 
Some efforts investigated privacy risks against pre-trained encoders, such as membership inference attacks~\cite{he2021quantifying, liu2021encodermi}, model extraction ~\cite{dziedzic2022difficulty, liu2022stolenencoder,sha2022can}.
At the same time, backdoor attacks and poisoning attacks, the common security threats that usually occur in the training phase, have been shown to be deleterious to pre-trained encoders~\cite{carlini2021poisoning,jia2022badencoder,liu2022poisonedencoder,saha2022backdoor}. 
In contrast, \textit{adversarial examples}, which appear during the testing phase and pose great threat against neural networks, have not been thoroughly investigated yet. 
A concurrent work, PAP~\cite{ban2022pre}, produced a pre-trained perturbation by lifting the feature activations of low-level layers, but the generated adversarial examples lack semantics and rely heavily on the pre-training dataset.
On the contrary, our work aims to achieve effective attacks from the perspective of directly changing the intrinsic texture features of the samples under more demanding conditions that better reflect 
realistic scenarios.

\subsection{Universal Adversarial Examples}
It is well known that deep neural networks are vulnerable to adversarial examples, where an attacker can fool the model by adding minor noise to the image, usually in the form of perturbation~\cite{carlini2017towards, goodfellow2014explaining, hu2022protecting, hu2023pointca} and patch \cite{hu2021advhash, liu2020bias, yang2020design}.
Universal adversarial attack \cite{moosavi2017universal} was proposed to fool the target model by imposing a single adversarial noise vector on all the images.
Existing works can be divided into optimization-based universal adversarial attacks~\cite{moosavi2017universal, mopuri2018generalizable, mopuri2017fast} and generative universal adversarial attacks~\cite{hayes2018learning, mopuri2018nag, mopuri2018ask}.
Compared with  optimization-based solution, generative universal adversarial attacks can generate more generalized and natural-looking adversarial examples by learning the distribution of samples.
However, existing generative universal adversarial attacks in supervised learning can only fool a single model and require the label  information of the model output. 
Since pre-trained encoders can only output the feature vector corresponding to the image, exiting attacks cannot be directly applied  to  the pre-trained encoders, let alone having no knowledge about the downstream tasks.
Some works also proposed different defenses against adversarial examples, such as data pre-processing, adversarial training~\cite{madry2017towards,tramer2019adversarial}, pruning~\cite{ zhu2017prune}, and fine-tuning~\cite{peng2022fingerprinting}. These methods can defend against adversarial samples at different phases.

\section{Methodology}

\subsection{Threat Model}
Following existing studies on attacking pre-trained encoders~\cite{jia2022badencoder,liu2022poisonedencoder,saha2022backdoor}, we assume the attacker has access to the pre-trained encoders (\eg, through purchasing or directly downloading from publicly available websites), but has no knowledge of the pre-training datasets and the downstream tasks. 
The goal of the attacker is  to conduct non-targeted adversarial attacks  to disable the downstream tasks or damage their accuracy. Specifically, the attacker uses the pre-trained encoder to design a downstream agnostic universal adversarial perturbation or patch  that applies to various kinds of the input images from different datasets.
Then the adversarial example can mislead all the downstream classifiers that inherit the victim pre-trained encoder.
We also assume that the downstream task undertaker (called user hereinafter) is able to fine-tune the linear layer or the pre-trained encoder for their cause and the model provider can adopt common defenses like adversarial training to purify the encoder.

\begin{figure}[!t]   
\centering
  \centering
    \subcaptionbox{Encoder-UAP}{\includegraphics[width=0.23\textwidth]{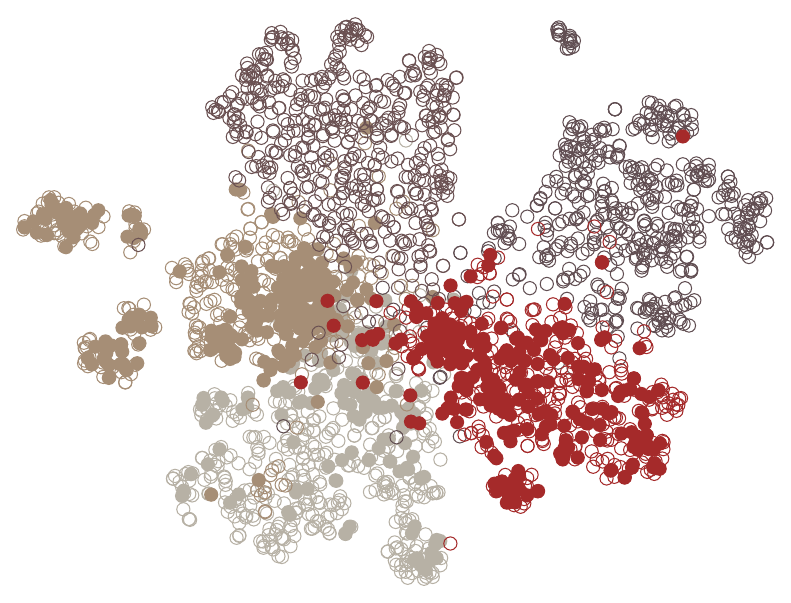}}
  \subcaptionbox{Downstream-UAP}{\includegraphics[width=0.23\textwidth]{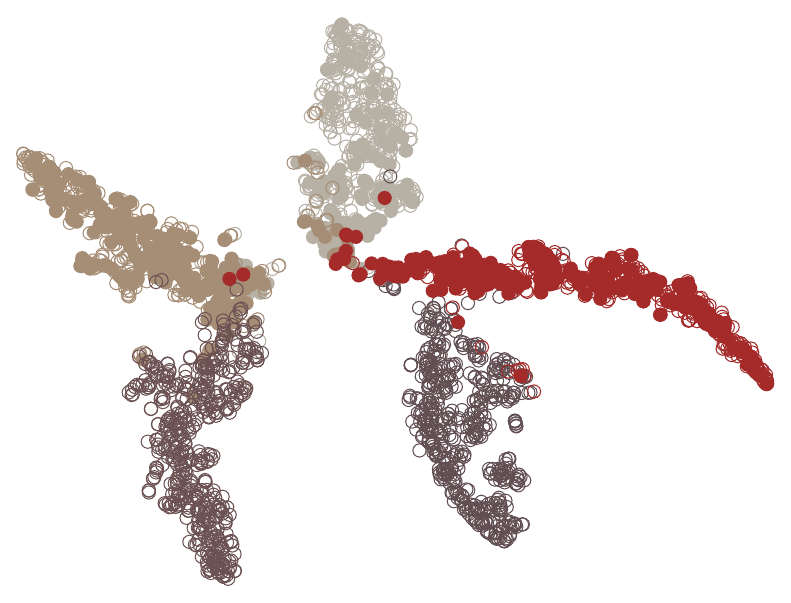}}
  \subcaptionbox{Encoder-AdvEncoder}{\includegraphics[width=0.23\textwidth]{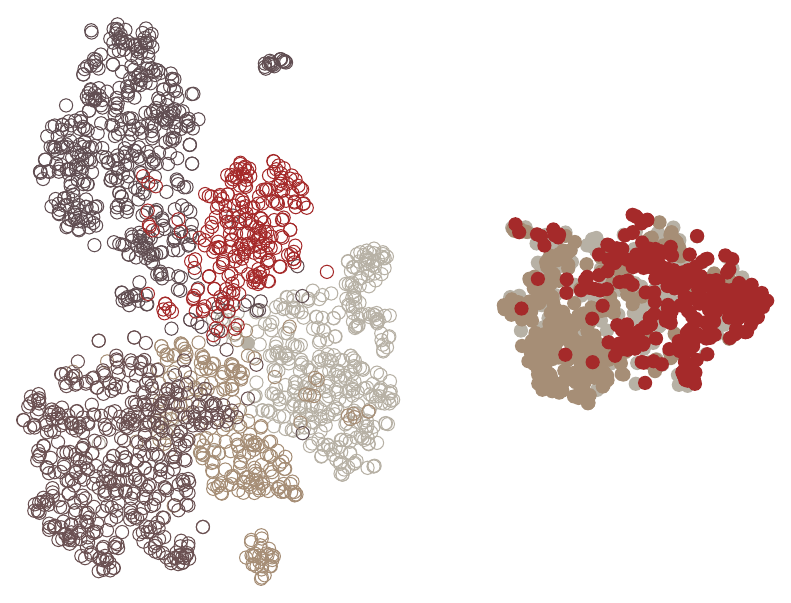}}
   \subcaptionbox{Downstream-AdvEncoder}{\includegraphics[width=0.23\textwidth]{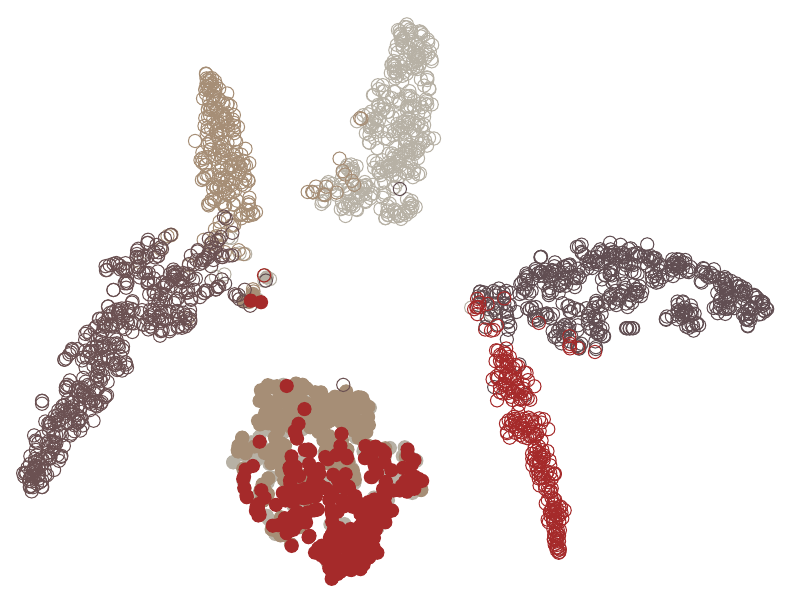}}
      \caption{t-SNE visualization of the feature space of the adversarial examples produced by UAP~\cite{moosavi2017universal} and AdvEncoder in the pre-trained encoder (based on SimCLR) and downstream model, respectively. Five different colors represent different classes. The hollow circles represent bengin examples, while the solid ones represent adversarial examples.}
       \label{fig:tsne}
       \vspace{-0.6cm}
\end{figure}

 \begin{figure*}[!t]

  \setlength{\belowcaptionskip}{-0.45cm}  
    \centering
    \includegraphics[scale=0.52]{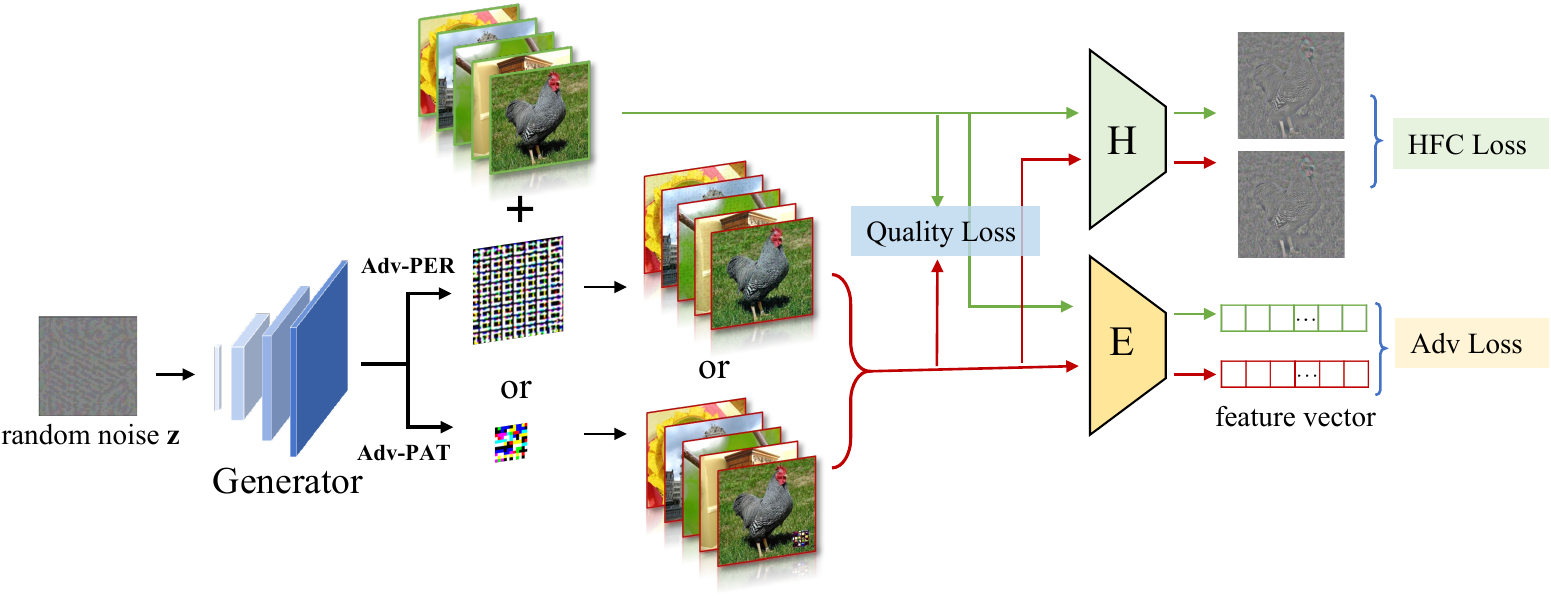}
    \caption{The pipeline of our attack}
    \label{fig:pipeline}
\end{figure*}

\subsection{Problem Definition}
Given an input $x \in \mathcal{D}_{p}$ to a pre-trained  encoder $ g_{\theta} (\cdot )$ that returns a feature vector $v \in \mathcal{V}$, a downstream classifier $f_{\theta^{'} } (\cdot )$ gives predictions based on the similarity of the feature vectors, where $\theta$ and $\theta^{'}$ denote the parameters of the pre-trained encoder and the downstream classifier, $x$ indicates any image in the dataset, $\mathcal{D}_{p}$ and $\mathcal{V}$ refer to the pre-training dataset and feature space, respectively.
The attacker uses an attacker's surrogate dataset $\mathcal{D}_{a}$, which is irrelevant to the pre-training dataset $\mathcal{D}_{p}$ and the downstream dataset  $\mathcal{D}_{d}$, to generate a universal adversarial noise against the pre-trained encoder. 
Additionally, the universal adversarial noise $\delta$ should be suffciently small, and modeled through an upper-bound $\epsilon$ on the $l_{p}$-norm.
This problem can be formulated as:
\begin{equation} \label{eq:1}
 g_{\theta}\left (  x + \delta  \right )  \ne g_{\theta}\left (  x  \right ), \quad  s.t.\left \| \delta  \right \| _{p}\le \epsilon
\end{equation}

The attacker's goal is to implement a universal non-target attack to fool the downstream classifier $f_{\theta^{'}}$.
When a universal adversarial noise $\delta$ is attached to the downstream dataset sample $x \in \mathcal{D}_{d}$ , it leads to misclassification. Therefore, the attacker's goal can be formalized as:
\begin{equation} \label{eq:2}
f_{\theta^{'} }( g_{\theta}\left (  x + \delta  \right )) \ne  f_{\theta^{'} }(g_{\theta}\left (  x  \right ) ), \quad s.t.  \left \| \delta  \right \|_{ p} \le \epsilon
\end{equation}

\subsection{Intuition Behind AdvEncoder}
The pre-trained encoder outputs similar feature vectors for similar images, which are close together in the feature space and far away from the images of other categories. 
The  downstream tasks will output decisions based on these feature vectors, thus the attacker needs to push the adversarial example away from its initial position as much as possible in the feature space.
In order to realize downstream-agnostic adversarial attack, there are two challenges ahead.

\textbf{Challenge I: Lack of supervised signals in pre-trained encoder.}
When the attacker feeds the image to the pre-trained encoder, it only obtains the corresponding feature vector instead of the label.
It is infeasible to effectively attack the pre-trained encoder with the traditional approaches of adversarial examples in supervised learning.
An intuitive idea is to add a large budget perturbation to the sample to make the pre-trained encoder misclassify it. 
However,  as seen from~\cref{fig:tsne}(a), a large budget perturbation will not necessarily achieve the above goal, but may simply be an internal movement within the same class, rather than in a direction away from that class.
Recent works~\cite{delac2005effects,madry2017towards,wang2020high} have revealed that surface-statistical content with high frequency property is essential for DNNs and adversarial perturbations also have this property. 
Therefore, we propose using a universal adversarial noise to change the high frequency component of the image, \ie, the texture information, to influence the output of the pre-trained encoder. It plays the role of  label guidance in the supervised learning, and it is easier to push the target samples out of the original decision boundaries from the perspective of directly altering the semantics of the image itself.

\textbf{Challenge II: Lack of information about the downstream tasks.}
In the pre-trained encoder to downstream task paradigm, where fine-tuning affects original feature boundaries of the model, the above approach that simply fools the pre-trained encoder can barely influence downstream task decisions.
As seen in~\cref{fig:tsne}(b), the adversarial examples that have left the original class are again correctly classified by the downstream model after the change of decision boundaries caused by fine-tuning.
We thus hope to make the adversarial examples far enough away from the original class by a universal adversarial noise under a small perturbation bound, as depicted in~\cref{fig:tsne}(c).
Consequently, the downstream classifier will be misled based on the apparent similarity of the feature vectors.
Given the remarkable capability of generative networks at generating features with fixed patterns, we further design a generative attack framework to improve the generalization of universal adversarial noise. As shown in~\cref{fig:tsne}(d),  all target samples will be clustered together in the feature space and get away from all the normal samples, making it difficult for downstream tasks to correctly classify target samples.

\subsection{Frequency-based Generative Attack Framework}
In this section, we present AdvEncoder, a novel generative attack against pre-trained encoder in self-supervised learning. 
The pipeline of AdvEncoder is depicted in~\cref{fig:pipeline}. It consists of an adversarial generator $\mathcal{G}$, a high frequency filter $\mathcal{H}$, and a victim encoder $\mathcal{E}$. 
Specifically, we design a frequency-based  generative attack framework to generate a universal adversarial noise. 
By feeding a fixed noise $z$ into the adversarial generator, we obtain a universal adversarial noise and paste it onto target image of the attacker's surrogate dataset $\mathcal{D}_{a}$ to get an adversarial example $x^{adv}$. 

The objective function of the adversarial generator $\mathcal{G}$ is:
\begin{equation}  \label{eq:3}
 \mathcal{L}_{\mathcal{G}} =\alpha  \mathcal{L}_{adv} + \beta  \mathcal{L}_{hfc} +\lambda \mathcal{L}_{q}
\end{equation}
where  $\mathcal{L}_{adv}$ is the adversarial loss function, $\mathcal{L}_{hfc}$  is the high frequency component loss function, $\mathcal{L}_{q}$ is the quality loss function, $\alpha$, $\beta$, $\lambda$ are pre-defined hyper-parameters.

$\mathcal{L}_{adv}$ enhances the attack strength of universal adversarial noise by maximizing the feature vector distance between the normal and adversarial samples of the encoder output.
We adopt InfoNCE \cite{oord2018representation} loss to measure the similarity between the output feature vectors of the pre-trained encoder $g (\cdot )$. Specifically, we treat the benign sample $x_{i} \in \mathcal{D}_{a}$ and the adversarial sample $x_{i}^{adv}$ as negative pairs, pulling away their feature distance. Thus $\mathcal{L}_{adv}$  is expressed as:

\begin{equation}  \label{eq:4}
\mathcal{L}_{adv} = log\left [ \frac{exp\left ( S\left ( g_{\theta}(x_{i}^{adv}), g_{\theta}(x_{i} ) \right )  \right /\tau ) }{ {\textstyle \sum_{j=0}^{K}} exp\left (  S\left ( g_{\theta}(x_{i}^{adv}), g_{\theta}(x_{j}) \right / \tau)  \right )  }  \right ]
\end{equation}
where 
$S \left ( \cdot \right ) $ denotes the cosine similarity measure function, $j$ is not equal to $i$, and $\tau$ indicates a temperature parameter.

Due to the lack of the guidance of label  information, pushing away the locations of output embeddings in the feature space  by adding noises alone requires large  perturbation budget.
$\mathcal{L}_{hfc}$ changes the original semantic features of the image by modifying the high frequency components
to further separate the  location of the target sample.

We can obtain the HFC of an image through the high frequency component filter $\mathcal{H}$.
The high frequency component loss $\mathcal{L}_{hfc}$  can be formalized as:

\begin{equation} \label{eq:5}
\mathcal{L}_{hfc} = - {\left \|\mathcal{H}(x^{adv})-\mathcal{H}(x)  \right \| } _{2}
\end{equation}

To achieve better stealthiness, we use $\mathcal{L}_{q}$ to control the magnitude of the adversarial noises output by the generator and crop $\delta$ after each optimisation to ensure it meets the constraints $\varepsilon$. Formally, we have:
\begin{equation} \label{eq:6}
\mathcal{L}_{q} = {\left \|x^{adv}- x  \right \| } _{2}
\end{equation}

Without changing the framework of AdvEncoder, we can convert a universal adversarial noise into two common forms of attacks, universal adversarial perturbation (AdvEncoder-Perturbation, abbreviated as Adv-PER) and universal adversarial patch (AdvEncoder-Patch, abbreviated as Adv-PAT).

\noindent\textbf{Adv-PER.} The attacker  directly adds the universal adversarial perturbation generated by the   generator  to the image, which has better stealthiness.
The perturbation-based adversarial example can be represented as:
\begin{equation} \label{eq:7}
x^{adv} = x +  \mathcal{G}(z)
\end{equation}

\noindent\textbf{Adv-PAT.} 
The attacker can apply the adversarial patch to the image with a randomly chosen hidden location to obtain the adversarial example. It is easier to be realized in the physical world.
The patch-based adversarial example can be represented as:
\begin{equation} \label{eq:8}
x^{adv}  = x  \odot (1- m) + \mathcal{G}(z)  \odot m
\end{equation}
where $ \odot$ denotes the element-wise product, $m$ is a binary matrix that contains the position information of the universal adversarial patch.

\section{Experiments}

\subsection{Experimental Setting}

 \noindent\textbf{Datasets and Models.} We use the publicly available pre-trained encoders from \emph{solo-learn}~\cite{JMLR:v23:21-1155}, an established SSL library, as victim encoders. 
 All the encoders are pre-trained on ImageNet~\cite{russakovsky2015imagenet} or CIFAR10~\cite{krizhevsky2009learning} with ResNet18 backbone. 
 For a comprehensive study, we  select fourteen SSL methods (Barlow Twins\cite{zbontar2021barlow}, BYOL\cite{grill2020bootstrap}, DeepCluster v2\cite{caron2020unsupervised}, DINO\cite{caron2021emerging},  MoCo v2+\cite{chen2020improved}, MoCo v3 \cite{chen2021empirical}, NNCLR\cite{dwibedi2021little}, ReSSL\cite{zheng2021ressl}, SimCLR \cite{chen2020simple}, SupCon\cite{khosla2020supervised}, SwAV\cite{caron2020unsupervised}, VIbCReg\cite{lee2021vibcreg}, VICReg\cite{bardes2021vicreg}, W-MSE\cite{ermolov2021whitening}). We make no strong assumptions about the attacker's knowledge, so we set the attacker's surrogate dataset to be CIFAR10 as the default setting. For different downstream tasks, we use the following four image datasets: STL10~\cite{coates2011analysis}, GTSRB~\cite{stallkamp2012man},  CIFAR10, and ImageNet.

\noindent\textbf{Evaluation Metrics.} We use \textit{Malicious Accuracy} (MA) and \textit{Attack Success Rate} (ASR) to evaluate the attack performance of our AdvEncoder. 
MA denotes the accuracy of adversarial examples being correctly classified, and ASR indicates the attack success rate.

 \begin{figure}[!t]
    \centering
    \includegraphics[scale=0.372]{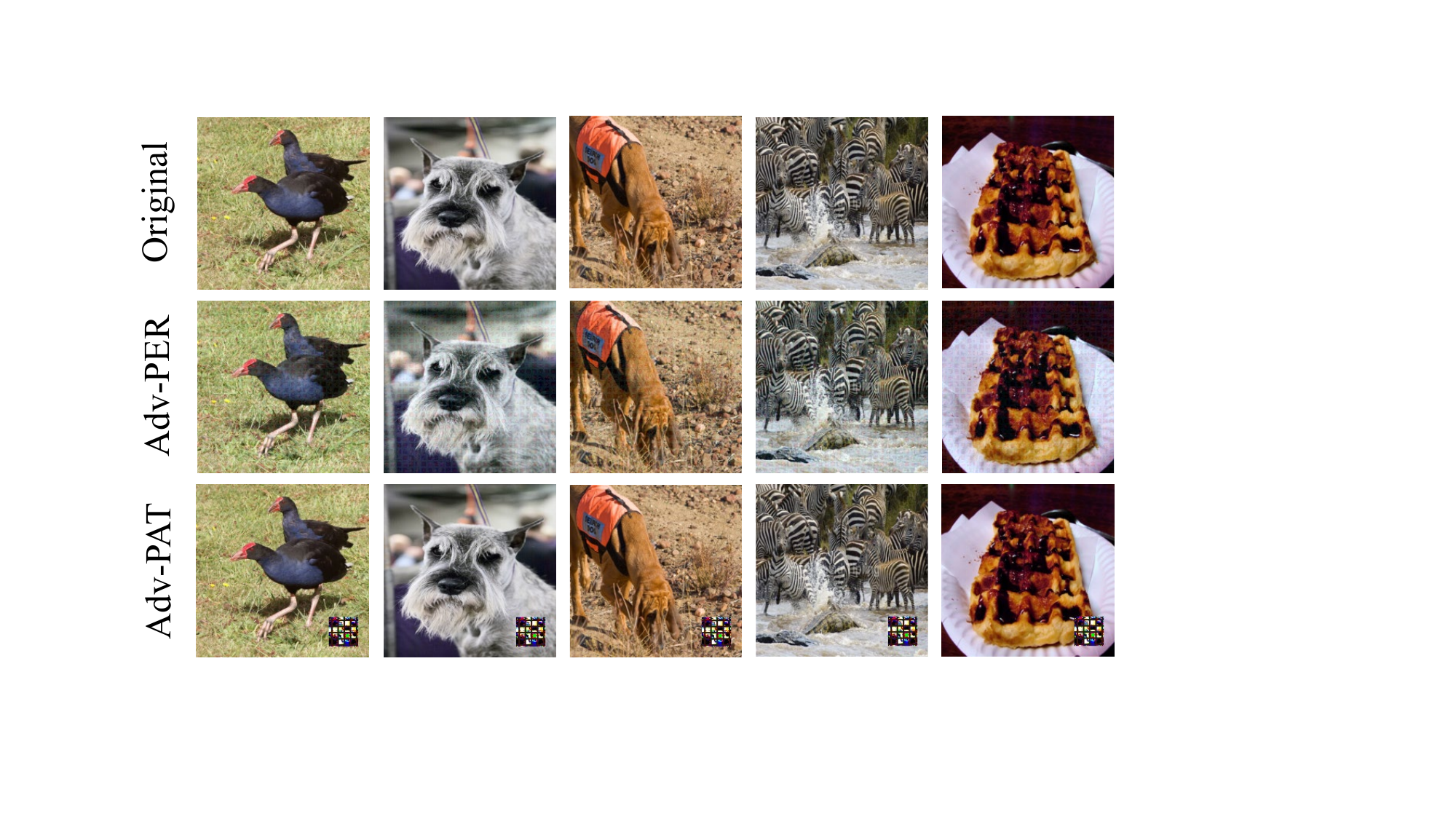}
    \caption{Adversarial examples generated by AdvEnocder based on ImageNet}
    \label{fig:demo}
    \vspace{-4mm}
\end{figure}

\begin{table*}[htbp]
\setlength{\abovecaptionskip}{2pt}
  \centering
  \caption{The ASR (\%) of Adv-PER under different settings.
  $\mathcal{S}_{1}$ - $\mathcal{S}_{4}$ denote the settings where the downstream datasets are CIFAR10, STL10, GTSRB, ImageNet, respectively, and all the attacker’s surrogate dataset is CIFAR10. $\mathcal{S}_{5}$ -  $\mathcal{S}_{8}$ use ImageNet as the attacker’s surrogate dataset, with the downstream datasets remained the same as $\mathcal{S}_{1}$ - $\mathcal{S}_{4}$. Barlow Twins and DeepCluster v2 are abbreviated as Barlow and DeepC2, respectively. }
  \scalebox{0.69}{
    \begin{tabular}{c|c|cccccccccccccc}
    \toprule[1.5pt]
    Dataset & Setting & Barlow & BYOL  & DeepC2   & DINO  & MoCo2+ & MoCo3 & NNCLR & ReSSL & SimCLR & SupCon & SwAV  & VIbCReg & VICReg & W-MSE \\
    \hline
     & $\mathcal{S}_{1}$    & 85.51  & 89.93  & 79.42  & 89.37  & 64.92  & \textbf{84.79}  & 88.48  & 87.85  & 57.07  & 93.62  & 87.71  & 87.15  & 90.04  & \textbf{89.26}  \\
   & $\mathcal{S}_{2}$    &  \textcolor[RGB]{169,169,169}{45.67}  & \textcolor[RGB]{169,169,169}{63.48}  &  \textcolor[RGB]{169,169,169}{58.25}  &  \textcolor[RGB]{169,169,169}{56.15}  &  \textcolor[RGB]{169,169,169}{32.08}  & 55.03  &  \textcolor[RGB]{169,169,169}{51.37}  &  \textcolor[RGB]{169,169,169}{51.60}  &  \textcolor[RGB]{169,169,169}{30.11}  &  \textcolor[RGB]{169,169,169}{71.88}  &  45.25  &  \textcolor[RGB]{169,169,169}{52.25}  &  \textcolor[RGB]{169,169,169}{68.38}  & \textcolor[RGB]{169,169,169}{55.10}  \\
    & $\mathcal{S}_{3}$    & 87.49  & 84.59  & 83.17  & 87.02  & \textbf{80.53}  & 80.40  & 91.09  & 92.73  & 69.30  & 91.26  & \textbf{92.11}  & 84.70  & 89.24  & 78.80  \\
          & $\mathcal{S}_{4}$    & 76.53  & 87.69  & 80.21  & 79.42  & 69.94  & 83.41  & 81.14  & 78.61  & 66.59  & 91.53  & 73.89  & 74.24  & 86.42  & 84.57  \\
       CIFAR10  & $\mathcal{S}_{5}$    & 90.46  & 85.94  & 87.42  & 89.99  & 58.00  & 76.44  & 90.10  & 88.23  & 72.20  & 89.43  & 72.28  & 89.41  & 89.03  & 78.87  \\
         & $\mathcal{S}_{6}$    & 85.85  & 74.93  & 88.96  & 70.64  & 33.51  &  \textcolor[RGB]{169,169,169}{43.35}  & 87.13  & 65.55  & 58.62  & 78.78  &  \textcolor[RGB]{169,169,169}{36.94}  & 80.44  & 74.89  & 60.68  \\
          & $\mathcal{S}_{7}$    & 97.19  & \textbf{95.52}  & 94.50  & \textbf{93.43}  & 79.59  & 82.98  & 91.59  & \textbf{93.45}  & \textbf{92.53}  & \textbf{95.96}  & 83.35  & 96.41  & \textbf{93.00}  & 73.07  \\
          & $\mathcal{S}_{8}$    & \textbf{97.42}  & 92.47  & \textbf{96.48}  & 90.31  & 69.60  & 75.19  & \textbf{96.30}  & 88.44  & 87.50  & 94.35  & 71.32  & \textbf{97.15}  & 90.93  & 82.74  \\
          & \textbf{AVG} & 83.26  & 84.32  & 83.55  & 82.04  & 61.02  & 72.70  & 84.65  & 80.81  & 66.74  & 88.35  & 70.36  & 82.72  & 85.24  & 75.39  \\

    \hline
     & $\mathcal{S}_{1}$    & 70.13  & 88.12  & \textbf{79.27}  & 83.29  & 82.33  & 72.52  & 70.91  & 87.86  & 71.94  & 76.37  & 84.00  & 82.77  & \textbf{82.51}  & 89.62  \\
          & $\mathcal{S}_{2}$    &  \textcolor[RGB]{169,169,169}{55.49}  &  \textcolor[RGB]{169,169,169}{58.67}  &  \textcolor[RGB]{169,169,169}{45.29}  &  \textcolor[RGB]{169,169,169}{61.67}  &  \textcolor[RGB]{169,169,169}{53.22}  &  \textcolor[RGB]{169,169,169}{59.51}  &  \textcolor[RGB]{169,169,169}{53.00}  &  \textcolor[RGB]{169,169,169}{59.63}  &  \textcolor[RGB]{169,169,169}{55.87}  &  \textcolor[RGB]{169,169,169}{48.73}  &  \textcolor[RGB]{169,169,169}{55.22}  &  \textcolor[RGB]{169,169,169}{61.01}  &  \textcolor[RGB]{169,169,169}{57.95}  &  \textcolor[RGB]{169,169,169}{67.96}  \\
           & $\mathcal{S}_{3}$    & 74.18  & 73.75  & 68.10  & 67.65  & 70.89  & 68.05  & 67.73  & 82.39  & 64.30  & 66.19  & 69.51  & 78.18  & 78.65  & 76.72  \\
           & $\mathcal{S}_{4}$    & 71.84  & 75.44  & 73.29  & 75.83  & 74.01  & 65.51  & 68.85  & 76.18  & 71.52  & 69.65  & 74.41  & 72.57  & 77.60  & 83.59  \\
         ImageNet & $\mathcal{S}_{5}$    & \textbf{87.94}  & \textbf{88.94}  & 77.28  & \textbf{83.97}  & \textbf{86.95}  & 76.11  & \textbf{86.32}  & \textbf{89.69}  & \textbf{88.95}  & \textbf{78.18}  & \textbf{86.54}  & \textbf{84.50}  & 81.64  & \textbf{90.61}  \\
         & $\mathcal{S}_{6}$    & 69.35  &  64.76  & 57.81  & 64.16  & 56.13  & 60.49  & 65.75  & 67.33  & 70.08  & 55.90  & 60.14  & 70.04  & 58.28  & 80.05  \\
         & $\mathcal{S}_{7}$    & 78.59  & 78.45  & 69.38  & 70.83  & 80.62  & \textbf{77.67}  & 74.05  & 86.13  & 83.70  & 69.05  & 81.17  & 81.65  & 79.76  & 85.56  \\
          & $\mathcal{S}_{8}$    & 80.02  & 80.28  & 77.48  & 77.52  & 76.74  & 75.72  & 74.73  & 81.36  & 79.68  & 71.01  & 80.20  & 80.33  & 78.32  & 90.03  \\
       & \textbf{AVG} & 73.44  & 76.05  & 68.49  & 73.12  & 72.61  & 69.45  & 70.17  & 78.82  & 73.26  & 66.88  & 73.90  & 76.38  & 74.34  & 83.02  \\
   \bottomrule[1.5pt]
    \end{tabular}%
}
  \label{tab:attack_performance_per}%
  
\end{table*}%

\begin{table*}[htbp]
\setlength{\abovecaptionskip}{2pt}
  \centering
  \caption{The ASR (\%) of Adv-PAT under different settings. $\mathcal{S}_{1}$ - $\mathcal{S}_{8}$ represent the same settings as mentioned in \cref{tab:attack_performance_per}.}
   \scalebox{0.69}{
    \begin{tabular}{c|c|cccccccccccccc}
    \toprule[1.5pt]
    Dataset & Setting & Barlow & BYOL  & DeepC2   & DINO  & MoCo2+ & MoCo3 & NNCLR & ReSSL & SimCLR & SupCon & SwAV  & VIbCReg & VICReg & W-MSE \\
    \hline
    & $\mathcal{S}_{1}$    & \textcolor[RGB]{169,169,169}{82.32}  & 88.20  & 90.88  & 81.77  &  \textcolor[RGB]{169,169,169}{81.52}  & 89.71  & 74.44  &  \textcolor[RGB]{169,169,169}{61.46}  & 89.87  & 69.19  & 89.31  & 63.32  & 82.15  & 89.13  \\
            & $\mathcal{S}_{2}$    & 88.16  &  \textcolor[RGB]{169,169,169}{80.08}  &  \textcolor[RGB]{169,169,169}{89.55}  & 77.95  & 84.03  &  \textcolor[RGB]{169,169,169}{82.10}  & 71.74  & 73.23  & 89.38  &  \textcolor[RGB]{169,169,169}{66.48}  &  \textcolor[RGB]{169,169,169}{85.60}  & 66.56  &  \textcolor[RGB]{169,169,169}{79.54}  & 82.56  \\
             & $\mathcal{S}_{3}$    & 93.89  & 92.02  & 94.43  & 89.98  & 98.40  & 90.21  & 89.84  & 89.15  & 96.22  & 91.19  & 97.09  & 89.88  & 90.65  & 89.22  \\
             & $\mathcal{S}_{4}$    & \textbf{97.86}  & 95.61  & 99.68  & 97.32  & 98.49  & 97.01  & 94.81  & 96.51  & \textbf{99.05}  & \textbf{96.81}  & \textbf{98.51}  & \textbf{96.25}  & 95.56  & \textbf{96.88}  \\
             CIFAR10 & $\mathcal{S}_{5}$    & 87.14  & 88.44  & 90.88  & 82.20  & 84.64  & 90.28  &  \textcolor[RGB]{169,169,169}{67.74}  & 66.53  & 89.90  & 76.34  & 89.31  & 62.79  & 84.68  & 89.25  \\
             & $\mathcal{S}_{6}$    & 88.00  & 86.12  & 89.71  &  \textcolor[RGB]{169,169,169}{76.61}  & 84.34  & 84.88  & 73.12  & 72.96  &  \textcolor[RGB]{169,169,169}{89.31}  & 67.63  & 86.84  &  \textcolor[RGB]{169,169,169}{56.24}  & 79.74  &  \textcolor[RGB]{169,169,169}{82.52}  \\
             & $\mathcal{S}_{7}$    & 93.91  & 91.76  & 94.69  & 87.15  & \textbf{99.20}  & 93.58  & 90.08  & 92.50  & 96.19  & 91.19  & 97.09  & 91.01  & 92.40  & 90.10  \\
             & $\mathcal{S}_{8}$    & 96.14  & \textbf{97.73}  & \textbf{99.69}  & \textbf{97.83}  & 98.40  & \textbf{98.44}  & \textbf{96.48}  & \textbf{98.11}  & \textbf{99.05}  & 96.27  & 98.03  & 95.28  & \textbf{96.65}  & 96.51  \\
           & \textbf{AVG}   & 90.93  & 89.99  & 93.69  & 86.35  & 91.13  & 90.78  & 82.28  & 81.31  & 93.62  & 81.89  & 92.72  & 77.67  & 87.67  & 89.52  \\
    \hline
    & $\mathcal{S}_{1}$    & 88.17  & 90.14  &  \textcolor[RGB]{169,169,169}{89.22}  & 89.41  &  \textcolor[RGB]{169,169,169}{89.90}  & 90.02  &  \textcolor[RGB]{169,169,169}{88.80}  & 92.01  & 90.30  & 90.50  &  \textcolor[RGB]{169,169,169}{90.06}  &  \textcolor[RGB]{169,169,169}{89.04}  & 89.49  & 91.21  \\
             & $\mathcal{S}_{2}$    &  \textcolor[RGB]{169,169,169}{82.35}  &  \textcolor[RGB]{169,169,169}{88.60}  & 89.98  &  \textcolor[RGB]{169,169,169}{89.07}  & 90.70  & 91.56  & 88.86  & 91.20  & 89.42  &  \textcolor[RGB]{169,169,169}{90.27}  & 90.48  & 89.66  & 85.14  &  \textcolor[RGB]{169,169,169}{89.86}  \\
            & $\mathcal{S}_{3}$    & 95.12  & \textbf{99.29}  & 96.89  & 94.67  & 94.01  & \textbf{98.49}  & 98.36  & 91.30  & 94.33  & 94.32  & 97.08  & 96.82  & 94.09  & 99.09  \\
             & $\mathcal{S}_{4}$    & \textbf{99.09}  & 98.18  & 99.16  & \textbf{98.79}  & \textbf{98.64}  & 98.29  & 98.34  & \textbf{98.51}  & \textbf{99.02}  & \textbf{99.10}  & 98.61  & \textbf{98.59}  & 98.56  & 98.51  \\
            ImageNet  & $\mathcal{S}_{5}$    & 89.00  & 90.14  &   \textcolor[RGB]{169,169,169}{89.22}  & 89.41  &  \textcolor[RGB]{169,169,169}{89.90}  &  \textcolor[RGB]{169,169,169}{88.83}  & 88.81  & 92.01  & 90.30  & 90.50  &  \textcolor[RGB]{169,169,169}{90.06}  &  \textcolor[RGB]{169,169,169}{89.04}  & 89.33  & 91.22  \\
             & $\mathcal{S}_{6}$    & 83.09  & 90.11  & 89.98  & 89.56  & 90.70  & 91.55  & 88.86  & 91.19  &  \textcolor[RGB]{169,169,169}{89.39}  & 90.70  & 90.38  & 90.78  &  \textcolor[RGB]{169,169,169}{84.42}  &  \textcolor[RGB]{169,169,169}{89.86}  \\
             & $\mathcal{S}_{7}$    & 95.37  & 95.64  & 97.38  & 93.18  & 91.19  & 98.20  & 98.36  &  \textcolor[RGB]{169,169,169}{90.19}  & 94.33  & 92.65  & 97.08  & 96.82  & 96.31  & \textbf{99.18}  \\
            & $\mathcal{S}_{8}$    & 98.89  & 98.19  & \textbf{99.21}  & 98.62  & \textbf{98.64}  & 98.47  & \textbf{98.70}  & 98.45  & \textbf{99.02} & 98.98  & \textbf{98.71}  & \textbf{98.59}  & \textbf{98.75}  & 98.47  \\
           & \textbf{AVG}   & 91.39  & 93.79  & 93.88  & 92.84  & 92.96  & 94.43  & 93.63  & 93.11  & 93.26  & 93.38  & 94.06  & 93.67  & 92.01  & 94.67  \\
    \bottomrule[1.5pt]
    \end{tabular}%
    }
  \label{tab:attack_performance_pat}%
\end{table*}%

\begin{table}[!h]
\setlength{\abovecaptionskip}{2pt}
  \centering
  \caption{Top-10 retrieval attack results. ``per-mAP'' represents the retrieval accuracy of the Adv-PER samples corresponding to clean samples, while ``pat-mAP'' denotes the accuracy of Adv-PAT samples.}
    \scalebox{0.65}{
    \begin{tabular}{c|c|c|c|c|c|c|c}
    \toprule[1.5pt]
      Dataset    & Metric & Barlow & BYOL  & DINO  & MoCo v2+ & NNCLR & SimCLR \\
     \hline
  \multirow{3}[2]{*}{STL10} & map   & 81.03 & 79.02 & 79.58 & 63.24 & 76.15 & 73.58 \\
          & per\_map & 23.26 & 21.76 & 22.95 & 22.77 & 24.99 & 21.76 \\
          & pat\_map & 21.15 & 19.64 & 21.12 & 26.89 & 26.22 & 23.59 \\
         \hline
    \multirow{3}[2]{*}{GTSRB} & map   & 93.92 & 85.68 & 88.49 & 84.37 & 83.83 & 88.53 \\
          & per\_map & 42.81 & 45.81 & 30.72 & 38.93 & 36.14 & 45.56 \\
          & pat\_map & 11.63 & 10.07 & 10.17 & 11.75 & 11.66 & 13.63 \\
    \bottomrule[1.5pt]
    \end{tabular}%
    }
  \label{tab:retrieval}%
   \vspace{-4mm}
\end{table}%

\subsection{Attack Performance }\label{sec:attack-performance}
\noindent\textbf{Implementation Details.} To demonstrate the effectiveness of AdvEncoder when the downstream task is unknowable, two types of downstream tasks, \emph{image classification} and \emph{image retrieval}, are chosen for testing.
Following~\cite{hu2021advhash,moosavi2017universal, mopuri2018nag}, we set $\epsilon$ (\ie, the perturbation budget of Adv-PER) to $10/255$ and the patch size (\ie, noise percentage of each sample) of  Adv-PAT to $0.03$.
We choose the bottom right corner of the image, which is not easily visible, to apply the patch. We set the hyper-parameters $\alpha= 1$, $\beta = 5$, $\lambda = 1$ and the training epoch to $20$ with batch size of $256$. 
The generator network is trained by Adam optimizer with the initial learning rate $0.0002$.

For the classification task, we attack fourteen types of SSL pre-trained encoders.
We evaluate AdvEncoder on each victim pre-trained encoder over four downstream tasks using two attacker's surrogate datasets, respectively.
As for the retrieval task, 
we attack six types of SSL encoders trained on CIFAR10 corresponding to the retrieval tasks of GTSRB and STL10.
We use mAP (\emph{mean average precision})~\cite{zuva2012evaluation}  to measure  the retrieval accuracy, and adapt per-mAP and pat-mAP to measure the retrieval accuracy for adversarial examples.
Adversarial examples generated by AdvEncoder are shown in~\cref{fig:demo}.

\noindent\textbf{Analysis.} Our experimental results on classification tasks reveal the severe vulnerability of downstream tasks based on pre-trained encoders.
Firstly, from \cref{tab:attack_performance_per} and \cref{tab:attack_performance_pat}, we can see that among the 224 attack settings, both Adv-PER and Adv-PAT perform well on all downstream tasks. In particular, Adv-PAT has a consistently high attack performance under different settings, with an average ASR of over 90\%.
Secondly, the attacker's surrogate dataset has an impact on the attack performance,  \eg, the ImageNet surrogate dataset outperforms CIFAR10. 
AdvEncoder performs better when the attacker's surrogate dataset is similar to the pre-training dataset and downstream dataset.
Thirdly, among the fourteen training methods, MoCo, SimCLR are more robust for adversarial examples, while BYOL, NNCLR, and SupCon are relatively weaker.
The  experimental results on image retrieval tasks in~\cref{tab:retrieval} also show that the adversarial examples generated by AdvEncoder can greatly affect the retrieval accuracy under different settings.

\begin{figure}[!t]   
\setlength{\abovecaptionskip}{4pt}
  \centering
     \subcaptionbox{Num-PER}{\includegraphics[width=0.15\textwidth]{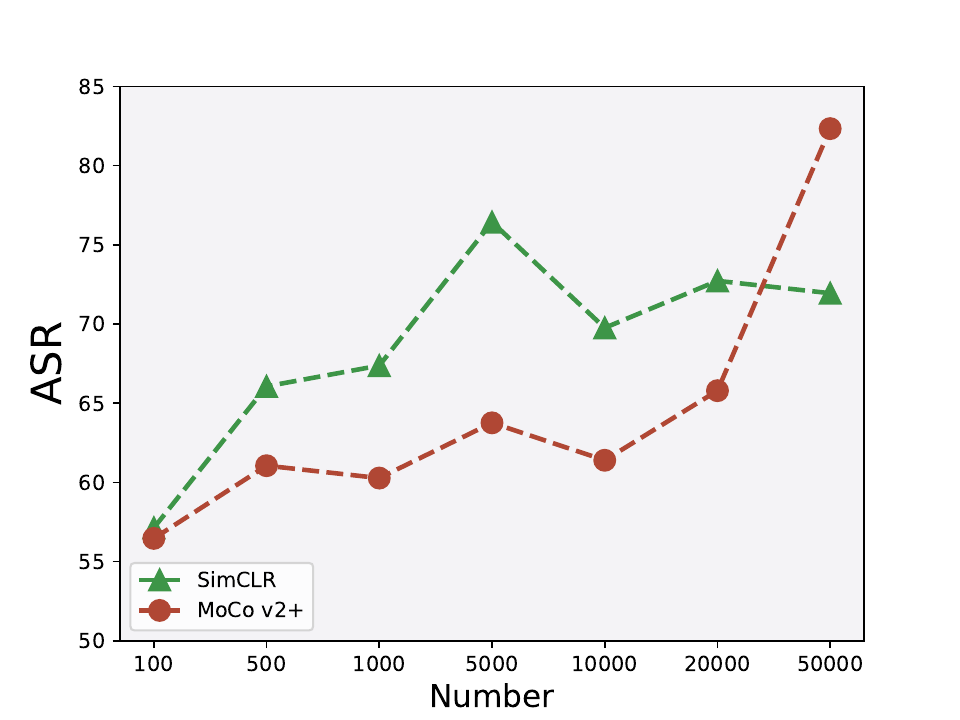}}
      \subcaptionbox{Num-PAT}{\includegraphics[width=0.15\textwidth]{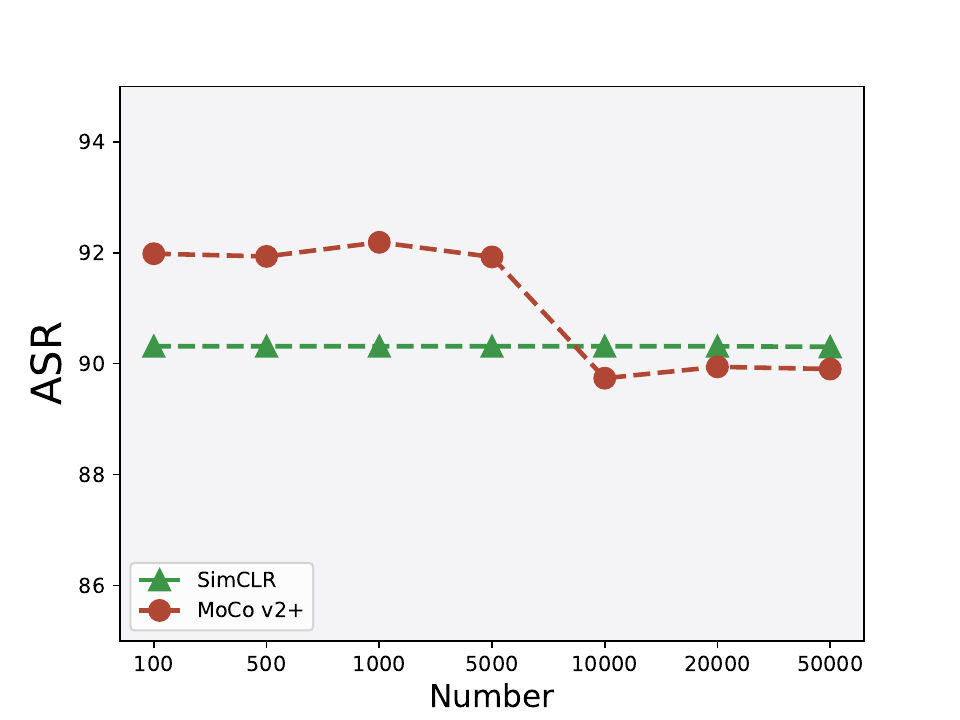}}
        \subcaptionbox{HFC-$\mathcal{G}$-PER}{\includegraphics[width=0.15\textwidth]{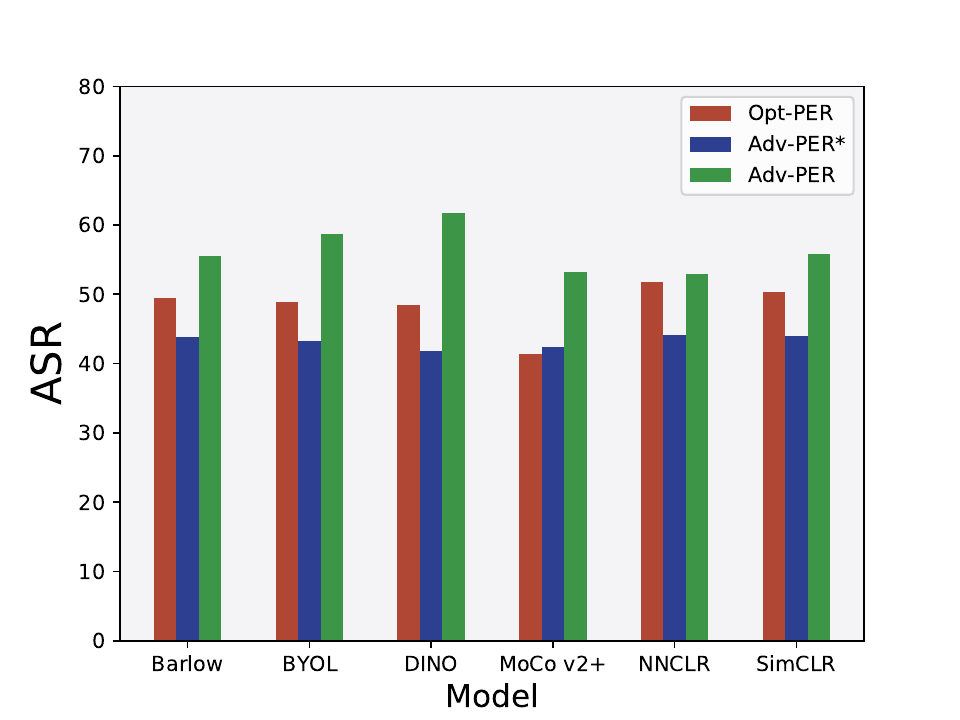}}
      \subcaptionbox{HFC-$\mathcal{G}$-PAT}{\includegraphics[width=0.15\textwidth]{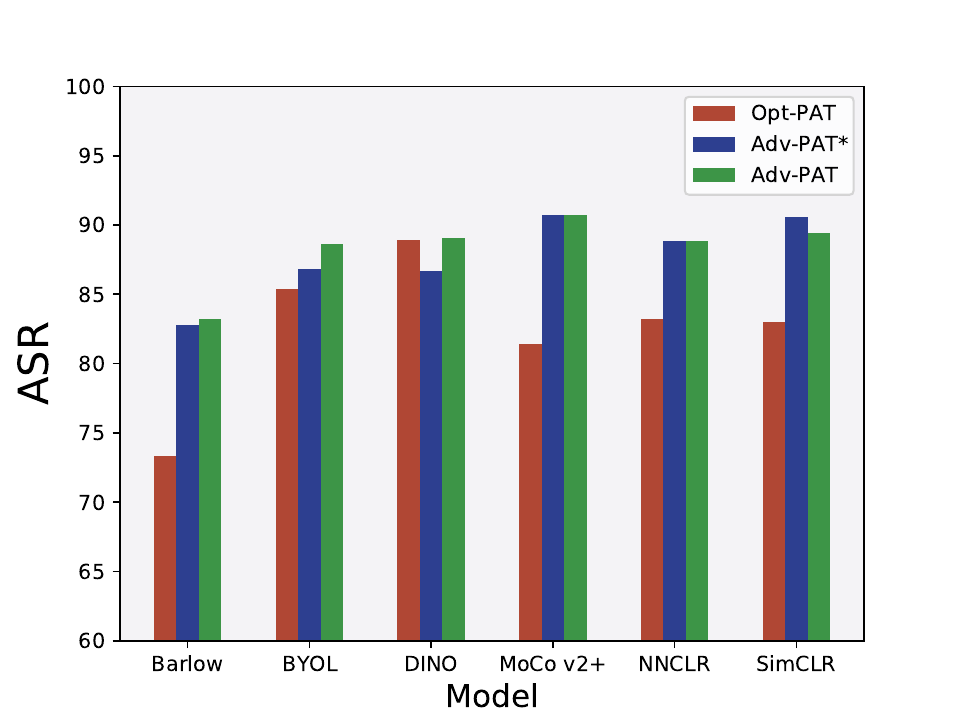}}
         \subcaptionbox{$\epsilon$}{\includegraphics[width=0.15\textwidth]{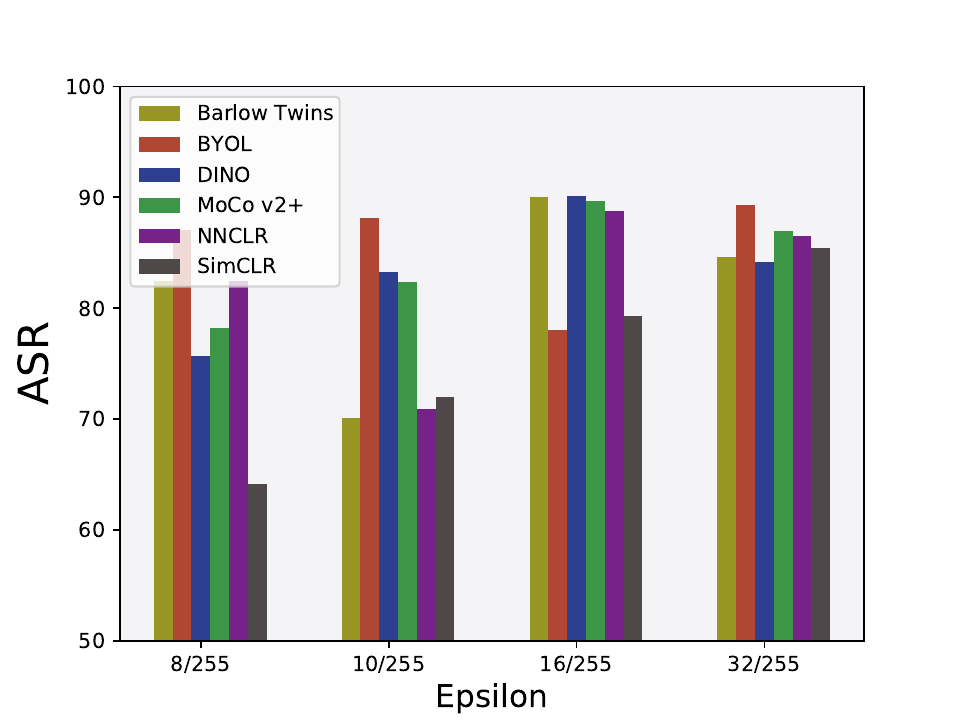}}
      \subcaptionbox{Patch-Size}{\includegraphics[width=0.15\textwidth]{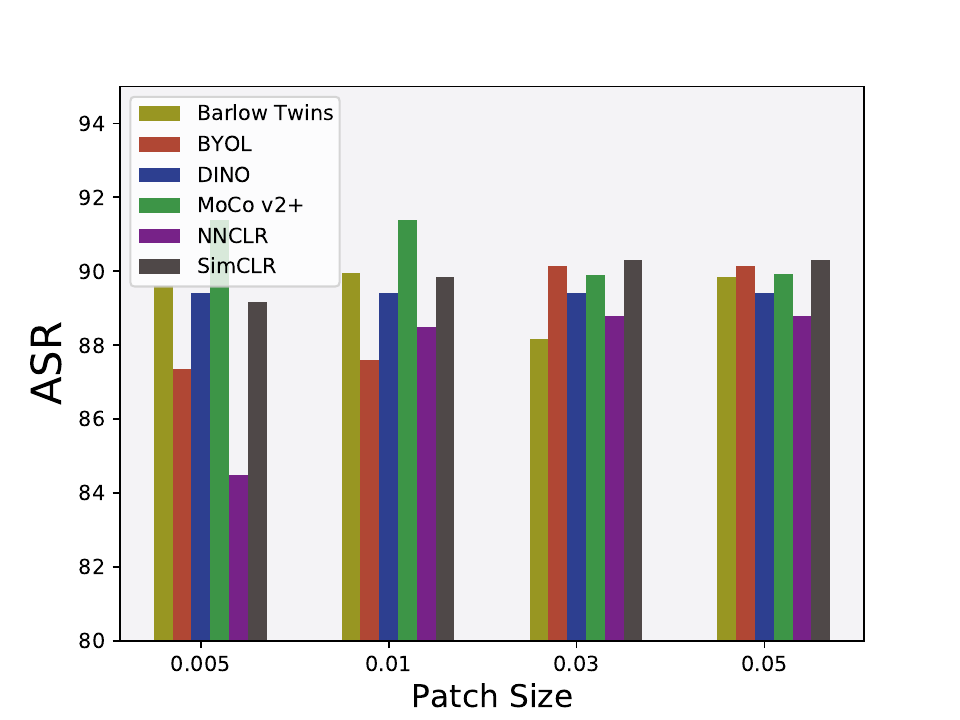}}
      \caption{The ablation study results. (a) - (d) examine the effect of the number of the surrogate data and different modules. (e) - (f) explore the effect of different attack strengths.}
       \label{fig:ablation results}
        \vspace{-0.6cm}
\end{figure}

\subsection{Ablation Study} \label{sec:ablation}
In this section, we explore the effect of different  attacker’s surrogate datasets, modules, and attack strengths on AdvEncoder.
We choose encoders trained on ImageNet and select CIFAR10 as the surrogate and downstream dataset.

\noindent\textbf{The Effect of Amount of Surrogate Data.} 
We investigate the effect of the limited sample size of the attacker's surrogate dataset.
Specifically, we randomly select different numbers of CIFAR10 samples to constitute the surrogate dataset and choose SimCLR and MoCo v2+ encoders for the attack.
The results in \cref{fig:ablation results}(a) - (b) show that the performance of Adv-PER generally improves with the increase of the number of samples.
For Adv-PAT, it performs well with different numbers of surrogate dataset settings.

\noindent\textbf{The Effect of HFC \& $\mathcal{G}$.} 
We analyze the effect of the HFC module and the generator module on the effectiveness of the scheme.
 We choose the downstream dataset as STL10.
In \cref{fig:ablation results}(c) - (d), Opt-PER and  Opt-PAT represent optimization-based versions of the same loss function, ``$*$'' denotes the version without HFC loss. Experimental results show that each module plays an important role.

\noindent\textbf{The Effect of $\epsilon$ \& Patch Size.} 
We study the effect of four different perturbation upper bound  $\epsilon$  and patch size on the attack performance of Adv-PER and Adv-PAT, respectively. 
From~\cref{fig:ablation results}(e), we can see that different pre-trained encoders have different sensitivities to different perturbation budgets.
The curves in~\cref{fig:ablation results}(f) show that the downstream tasks are more vulnerable to adversarial patches.

\begin{figure}[!t]   
\setlength{\abovecaptionskip}{4pt}
  \centering
         \subcaptionbox{CIFAR10-ImageNet}{\includegraphics[width=0.224\textwidth]{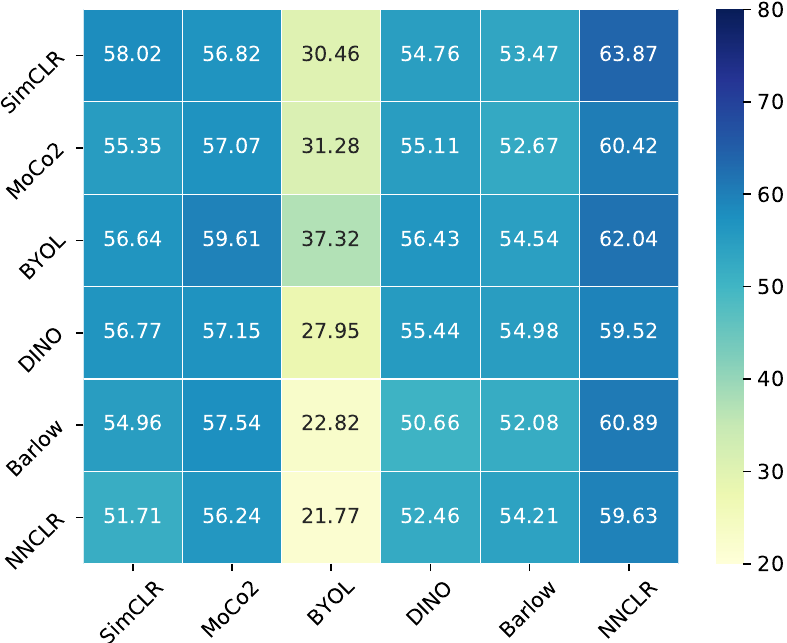}}
    \subcaptionbox{ImageNet-ImageNet}{\includegraphics[width=0.224\textwidth]{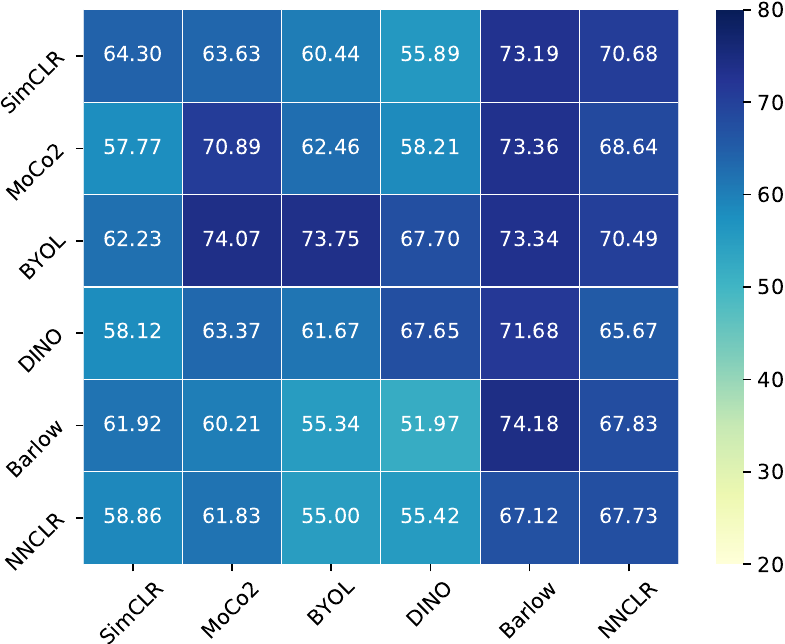}}
      \caption{The results (\%) of transferability study}
       \label{fig:trans}
        \vspace{-0.6cm}
\end{figure}

 \subsection{Transferability Study}

In this section, we choose Adv-PER as a representative to analyze the transferability of our scheme from two perspectives, namely, crossing pre-training datasets and SSL methods. To this end, we conduct experiments using CIFAR10 as the attacker's surrogate dataset and GTSRB as the downstream dataset. 
In~\cref{fig:trans}(a) - (b), CIFAR10-ImageNet represents we use CIFAR10 and ImageNet to train two encoders based on which adversarial examples and downstream tasks are made, respectively. ImageNet-ImageNet has the same definition. 
Each column represents different downstream models attacked with the same adversarial examples.
The results indicate that the Adv-PER method can effectively transfer attacks across different pre-training datasets and SSL methods, even without any prior knowledge of the pre-training and downstream datasets.

\begin{table}[htbp]
  \centering
  \caption{The ASR (\%) of comparison study}
  \scalebox{0.69}{
    \begin{tabular}{c|c|c|c|c|c|c}
    \toprule[1.5pt]
    Method & Barlow & BYOL  & DINO  & MoCo v2+ & NNCLR & SimCLR \\
    \hline
    UAP~\cite{moosavi2017universal}   & 48.95  & 45.97  & 43.01  & 42.24  & 46.41  & 48.41  \\
    UPGD~\cite{deng2020universal} & 18.31  & 23.89  & \textcolor[RGB]{169,169,169}{18.59}  & 18.31  & \textcolor[RGB]{169,169,169}{17.57}  & 20.83  \\
    FFF~\cite{mopuri2017fast}  & 47.71  & 50.26  & 45.53  & 50.33  & 46.64  & 52.33 \\
    SSP~\cite{naseer2020self}   & 50.40  & 46.30  & 47.12  & 50.87  & 49.51  & 48.06  \\
    NAG~\cite{mopuri2018nag}   & \textcolor[RGB]{169,169,169}{8.98}  & \textcolor[RGB]{169,169,169}{10.91}  & 26.22  & \textcolor[RGB]{169,169,169}{14.34}  & 24.49  & \textcolor[RGB]{169,169,169}{6.76}  \\
    PAP-base~\cite{ban2022pre}   & 37.89  & 27.29  & 25.52  & 30.47  & 24.65  & 46.89 \\
    PAP-fuse~\cite{ban2022pre}   & 46.61  & 45.47  & 30.26  & 45.14  & 40.05  & 46.30 \\
    PAP-ugs~\cite{ban2022pre}   & 38.20  & 36.89  & 30.55  & 37.71  & 42.99  & 53.56  \\
    Adv-PER  & \textbf{55.49}  & \textbf{58.67}  & \textbf{61.67}  & \textbf{53.22}  & \textbf{53.00}  & \textbf{55.87}  \\
    \hline
    UA-PAT~\cite{brown2017adversarial} & \textcolor[RGB]{169,169,169}{58.20}  & \textcolor[RGB]{169,169,169}{34.62}  & \textcolor[RGB]{169,169,169}{46.12}  & \textcolor[RGB]{169,169,169}{79.04}  & \textcolor[RGB]{169,169,169}{49.73}  & \textcolor[RGB]{169,169,169}{50.35}  \\
    Adv-PAT & 82.35  & \textbf{88.60}  & \textbf{89.07}  & \textbf{90.70}  & \textbf{88.86}  & \textbf{89.42}  \\
    \bottomrule[1.5pt]
    \end{tabular}
}
  \label{tab:comparision}%

\end{table}%

\subsection{Comparison Study}
\noindent\textbf{Implementation Details.} 
In this section, 
we compare Adv-Encoder with state-of-the-art adversarial attacks. 
PAP~\cite{ban2022pre} is the most similar work  with ours as it produces pre-trained perturbations from the perspective of model feature activation values. In contrast, we address the  attack inheritance of adversarial samples by directly changing the important 
 texture features of the samples themselves.
To further demonstrate our superiority, for perturbation we compare with classic optimized-based UAP schemes (\eg, UAP~\cite{moosavi2017universal}, UPGD~\cite{deng2020universal}, FFF~\cite{mopuri2017fast}, SSP~\cite{naseer2020self}, and PAP~\cite{ban2022pre}) and generative-based UAP scheme (\eg,  NAG~\cite{mopuri2018nag}).
For patch, we compare Adv-PAT with UA-PAT~\cite{brown2017adversarial}, an optimization-based adversarial patch method, which maintains the same experimental setup as the Adv-Encoder. 
Since the above supervised learning adversarial attacks can not be directly applied to attack the pre-trained encoder, 
we enable those UAP schemes to have complete pre-trained model (i.e., encoder connected with classification head trained under the same pre-training dataset).
For a comprehensive comparison under the pre-trained encoder to downstream task paradigm, we choose six encoders trained on ImageNet with CIFAR10 for the attacker’s surrogate dataset and STL10 for the downstream dataset.

\noindent\textbf{Analysis.} 
From \cref{tab:comparision}, we can see that AdvEncoder outperforms the other solutions without knowing the pre-training dataset and the downstream dataset.
Adv-PER shows superior performance compared to optimization-based and generative-based methods. Furthermore, AdvEncoder achieves better overall attack performance than the most relevant existing work, PAP, across six pre-trained encoders. Notably, Adv-PAT outperforms UA-PAT with an average ASR of over 85\%. Importantly, our method achieves these results without requiring additional classification headers, and instead directly leverages the pre-trained encoders for the attack.

\begin{figure}[!t]   
\setlength{\abovecaptionskip}{4pt}
  \centering
        \subcaptionbox{Corruption}{\includegraphics[width=0.15\textwidth]{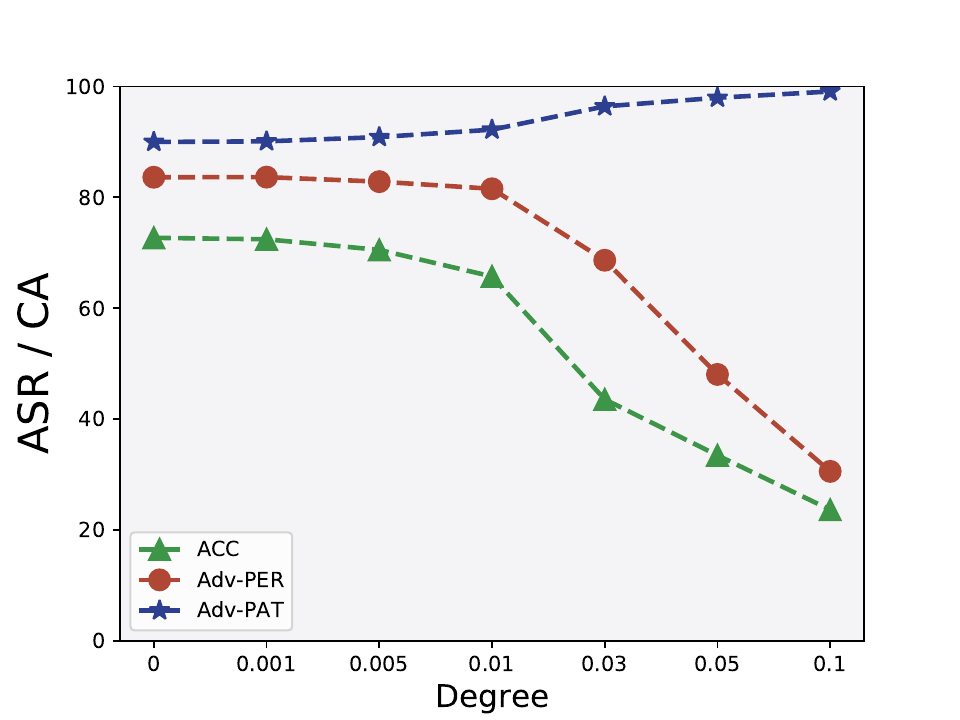}}
    \subcaptionbox{Fine-tune-PER}{\includegraphics[width=0.15\textwidth]{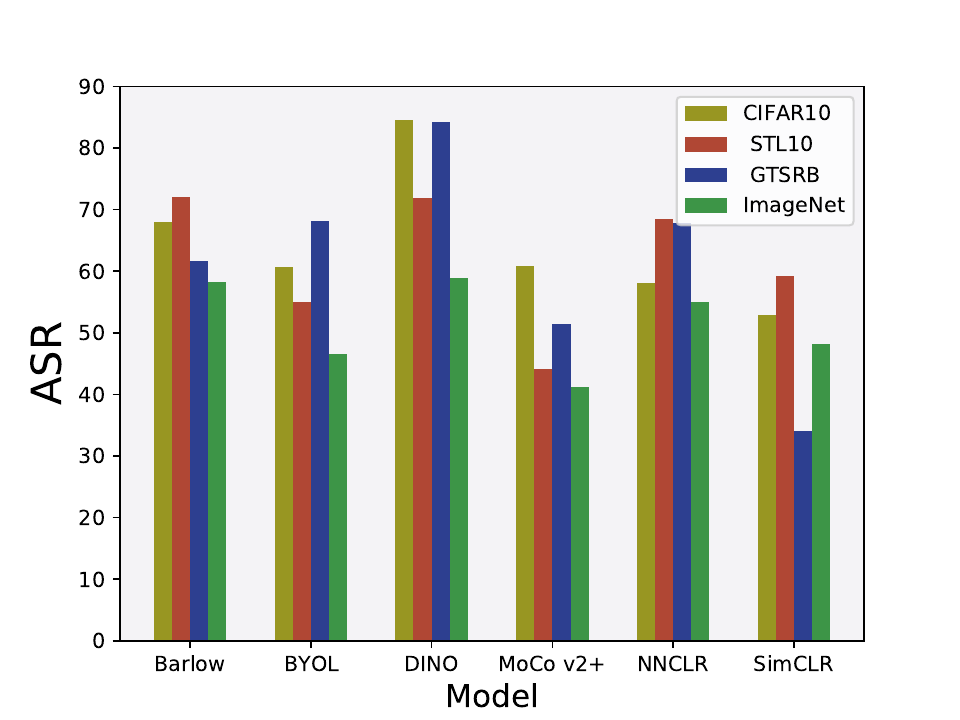}}
      \subcaptionbox{Fine-tune-PAT}{\includegraphics[width=0.15\textwidth]{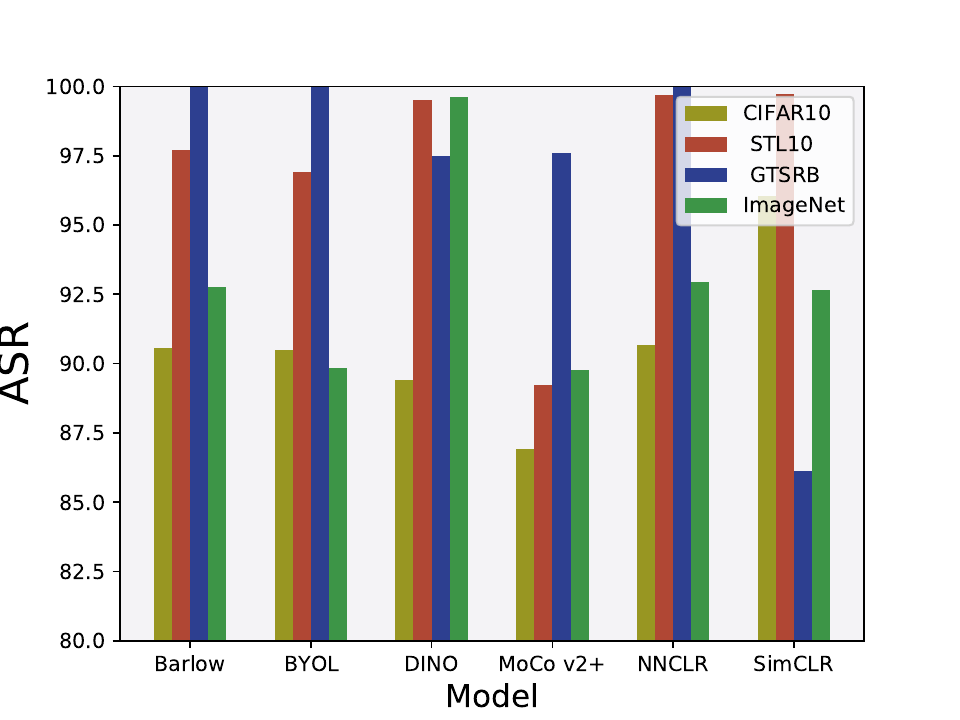}}
        \subcaptionbox{Pruning}{\includegraphics[width=0.15\textwidth]{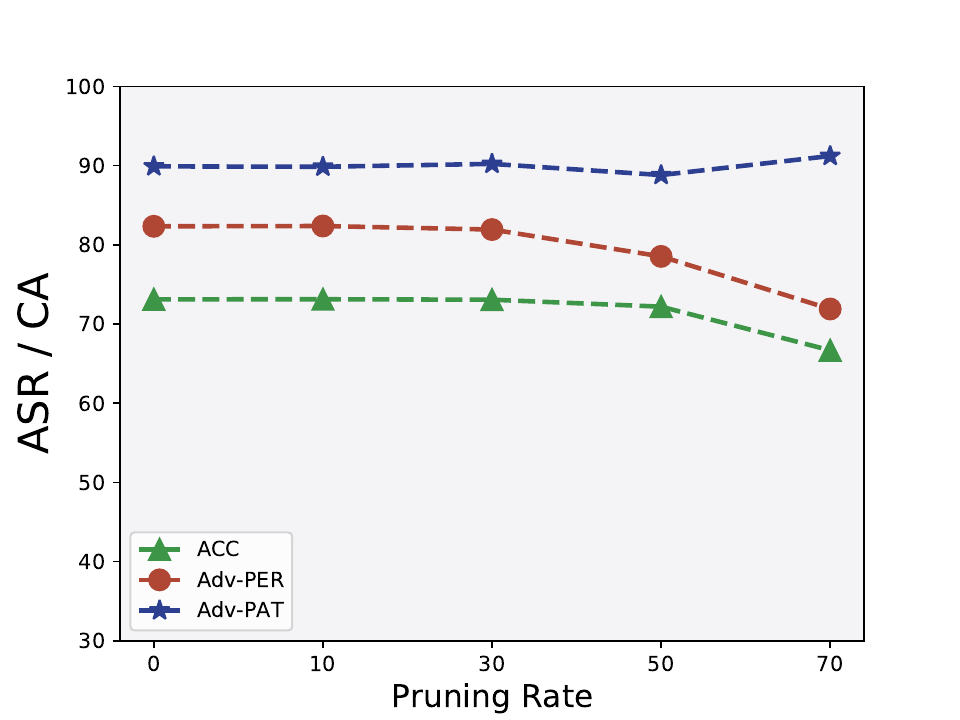}}
        \subcaptionbox{AT-PER}{\includegraphics[width=0.15\textwidth]{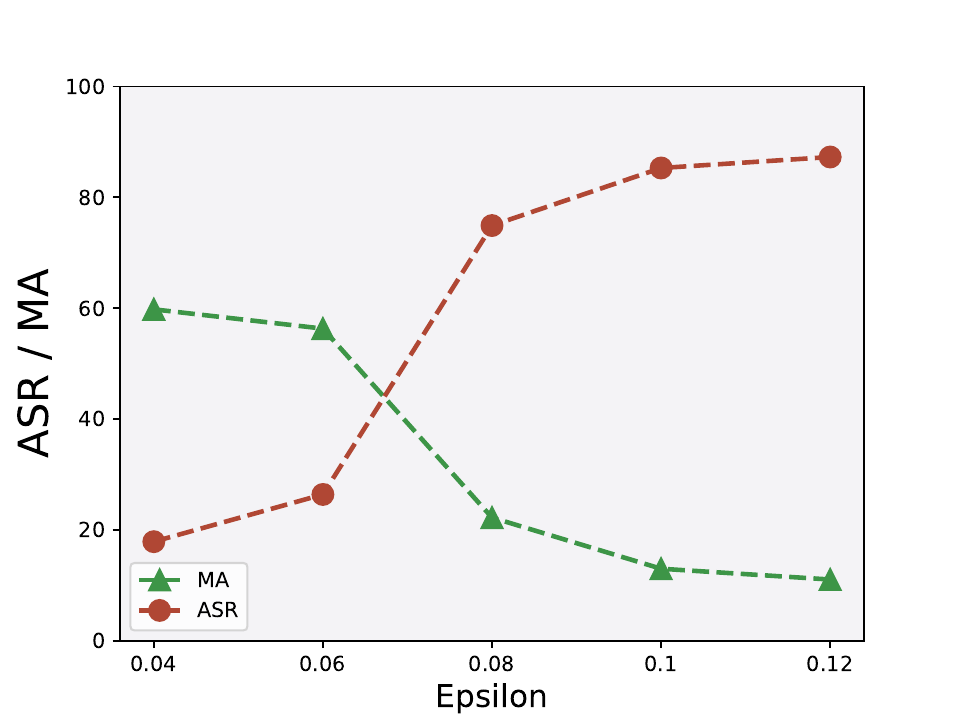}}
      \subcaptionbox{AT-PAT}{\includegraphics[width=0.15\textwidth]{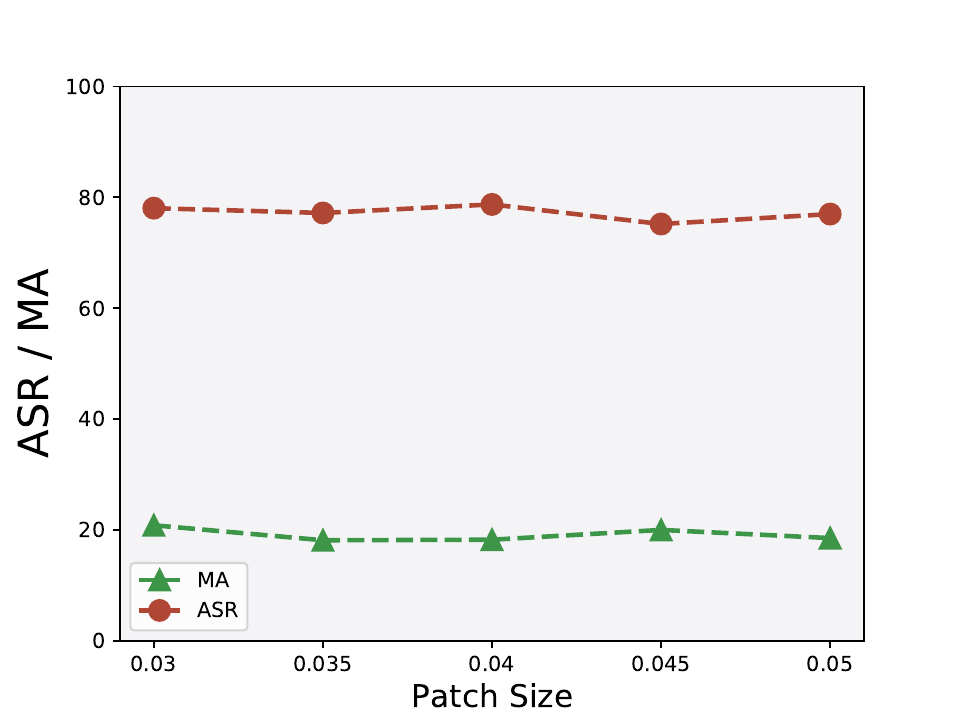}}
      \caption{The attack performance (\%) of AdvEncoder in different settings. (a) - (d) examine  three defenses Corruption, Fine-tuning, and Pruning on CIFAR10. (e) - (f) show the effect of adversarial training of the pre-trained encoder (based on SimCLR) on AdvEncoder.}
       \label{fig:defense}
        \vspace{-0.6cm}
\end{figure}

\section{Defense}
In this section, we tailor four defensive measures trying to mitigate AdvEncoder from the perspective of the user and the model provider, respectively.
For users using pre-trained encoders, we adopt pre-processing the input data, fine-tuning the entire model using a small amount of data, and pruning the parameters to defend against adversarial attacks.
As for the model providers, the defender can perform adversarial training on pre-trained encoders.
In the following experiments, we use the default settings from \cref{sec:ablation}. 

\subsection{Corruption}
We defend against adversarial examples by corrupting the images through adding different degree of Gaussian noise to the samples. As illustrated in ~\cref{fig:defense}(a), 
the accuracy of the model decreases significantly as the degree of Gaussian noise added increases. 
In particular, Adv-PER only experiences a slight decrease in accuracy when the degree of Gaussian noise is increased to 0.03, while Adv-PAT is almost unaffected. 
These findings indicate that Adv-Encoder can effectively resist the corruption-based pre-processing defense.

\subsection{Fine-tuning \& Pruning}
Fine-tuning~\cite{peng2022fingerprinting} and pruning~\cite{zhu2017prune} are two commonly used methods for downstream models to inherit pre-trained encoders, 
providing better adaptability to downstream tasks.
We first  
fully fine-tune pre-trained encoders based on MoCo v2+, using ten classes in CIFAR10, STL10, GSTRB, and ImageNet, respectively.
The results in ~\cref{fig:defense}(b) - (c) show that AdvEncoder still has excellent attack performance even after the encoder is fully fine-tuned.
Furthermore, 
we choose pruning rate in  $[0.1, 0.7]$, the results in~\cref{fig:defense}(d) show that AdvEncoder is able to resist the defenses based on model parameter pruning.

\subsection{Adversarial Training}
Adversarial training improves the robustness of the pre-trained encoder and poses a greater challenge to the adversarial examples.
Following~\cite{jiang2020robust}, we use the ImageNet dataset for adversarial training of the pre-trained encoder and choose CIFAR10 as downstream dateset. 
As demonstrated  in~\cref{fig:defense}(e) - (f),
we explore the degree of resistance to adversarial training for AdvEncoder of different attack strengths.
Adversarial training  slightly affects Adv-PER, but our attack can succeed after improving the attack strength. And Adv-PAT has not been affected at all.

\section{Conclusion}
In this paper, we propose the first generative attack to construct downstream-agnostic adversarial examples in self-supervised learning. 
It is a flexible framework that can generate both universal adversarial perturbations and patches.
We verify the excellent attack performance of Adv-Encoder on four downstream tasks corresponding to fourteen publicly available SSL encoders over two pre-training datasets.
We tailor four popular defenses to mitigate Adv-Encoder. The results further prove the attack ability of AdvEncoder and highlight the needs of new defense mechanism to defend pre-trained encoders.

\noindent\textbf{Acknowledgments.} Shengshan's work is supported in part by the National Natural Science Foundation of China (Grant No.U20A20177) and Hubei Province Key R\&D Technology Special Innovation Project under Grant No.2021BAA032. 
Qian's work is supported in part by the National Natural Science Foundation of China under Grants U20B2049 and U21B2018.
Shengshan Hu is the corresponding author.

{\small
\bibliographystyle{ieee_fullname}
\bibliography{egbib}
}

\appendix
\section{Appendix Contents}
\setcounter{table}{0}
\setcounter{figure}{0}
\renewcommand{\thetable}{A\arabic{table}}
\renewcommand{\thefigure}{A\arabic{figure}}

\begin{itemize}

\item Sec.~\ref{dataset}: 
Details of the datasets. 
\item Sec.~\ref{optimization}: 
Optimization procedure of AdvEncoder. 
We describe the optimization of our attack in detail.

\item Sec.~\ref{attack_performance}: 
Supplemental results of AdvEncoder's attack performance on \textbf{Image Classification \& Retrieval}, \textbf{Object Detection}, and  \textbf{Semantic Segmentation} tasks.

\item Sec.~\ref{ablation}: 
Supplemental ablation study about the effect of different backbones and random seeds on AdvEncoder.

\item Sec.~\ref{transferability}: 
Supplemental transferability results.

\item Sec.~\ref{Visualization}: 
Visualization of perturbations/patches generated by AdvEncoder.
\end{itemize}

\section{Datasets}\label{dataset}

Our experiments are based on the following four datasets: CIFAR10~\cite{krizhevsky2009learning}, STL10~\cite{coates2011analysis}, GTSRB~\cite{stallkamp2012man}, and ImageNet~\cite{russakovsky2015imagenet}. 
Specifically, we use the above four datasets as the attacker’s surrogate datasets and the downstream datasets, respectively.
Following~\cite{jia2022badencoder}, we resize all examples to 64x64x3. The details of the datasets are as follows:

\noindent\textbf{CIFAR10.} CIFAR10 contains 50,000 training images and 10,000 testing images. Each image has a size of 32×32×3 and belongs to one of 10 classes. 

\noindent\textbf{STL10.} STL10 contains 5,000 labeled training images and 8,000 labeled testing images, each of which has a size of 96×96×3. Moreover, the dataset contains 10 classes and each image belongs to one of them.  

\noindent\textbf{GTSRB.} GTSRB contains 51,800 traffic sign images in 43 categories. Each image has a size of 32×32×3. The dataset is divided into 39,200 training images and 12,600 testing images. 

\noindent\textbf{ImageNet.} ImageNet contains $1.2M$ training samples and $50,000$ testing samples with $1000$ classes. Each image has a size of 256×256×3. We randomly select $100$ classes from ImageNet to build our dataset.

\section{Optimization}\label{optimization}
In this section, we describe the optimization process of AdvEncoder in detail.
Given a random noise $z$, the generator generates a universal adversarial noise $\delta$ of the same size as the input image. 
We need to first crop the universal adversarial noise $\delta$ to within the imperceptibility constraint $\epsilon$. Then the universal adversarial noise can be converted into two forms of adversarial perturbation and adversarial patch by using \cref{eq:7} and \cref{eq:8}, respectively. 
By minimizing the objective function mentioned in \cref{eq:3}, we can optimize the generator to generate more generalized and transferable universal adversarial  perturbations or patches.
The whole optimization process is outlined in Algorithm.~\ref{optimization_advencoder}.

\begin{table}[htbp]
\setlength{\abovecaptionskip}{2pt}
  \centering
   \caption{The clean retrieval accuracy (\%) of different downstream models based on pre-trained encoders over two datasets. PD denotes the pre-training dataset and DD represents the downstream dataset.}
      \scalebox{0.6375}{
    \begin{tabular}{c|c|c|c|c|c|c|c|c}
    \toprule[1.5pt]
    PD & DD & Model & top1  & top5  & top10 & top20 & top50 & top100 \\
    \hline
    \multirow{12}[4]{*}{CIFAR10} & \multirow{6}[2]{*}{STL10} & Barlow & 99.33  & 93.31  & 80.22  & 60.36  & 36.86  & 24.67 \\
          &       & BYOL  & 99.00  & 93.32  & 81.33  & 63.17  & 36.82  & 25.25 \\
          &       & DINO  & 100.00  & 93.62  & 82.79  & 64.45  & 38.72  & 25.67 \\
          &       & MoCo2+  & 99.67  & 93.74  & 82.74  & 63.56  & 37.21  & 25.4 \\
          &       & NNCLR & 100.00  & 94.21  & 82.83  & 63.99  & 36.34  & 25.01 \\
          &       & SimCLR & 99.67  & 94.45  & 80.55  & 62.16  & 38.06  & 25.28 \\
\cline{2-9}          & \multirow{6}[2]{*}{GTSRB} & Barlow & 96.67  & 97.43  & 96.39  & 94.52  & 92.38  & 90.42 \\
          &       & BYOL  & 95.33  & 94.65  & 92.24  & 89.66  & 85.99  & 82.54 \\
          &       & DINO  & 94.33  & 94.86  & 92.80  & 89.51  & 84.63  & 80.41 \\
          &       & MoCo2+  & 85.00  & 87.19  & 86.17  & 84.45  & 81.72  & 78.62 \\
          &       & NNCLR & 90.33  & 91.43  & 89.58  & 87.57  & 82.85  & 79.92 \\
          &       & SimCLR & 94.00  & 94.19  & 92.30  & 90.58  & 87.73  & 85.43 \\
    \hline
    \multirow{12}[4]{*}{ImageNet} & \multirow{6}[2]{*}{STL10} & Barlow & 99.67  & 94.07  & 81.03  & 62.26  & 36.67  & 25.02 \\
          &       & BYOL  & 93.33  & 90.53  & 79.02  & 61.63  & 37.58  & 24.92 \\
          &       & DINO  & 94.00  & 91.38  & 79.58  & 61.00  & 36.33  & 24.6 \\
          &       & MoCo2+  & 65.67  & 70.93  & 63.24  & 49.50  & 32.28  & 23.38 \\
          &       & NNCLR & 86.67  & 86.63  & 76.15  & 59.27  & 35.72  & 24.67 \\
          &       & SimCLR & 83.33  & 84.48  & 73.58  & 56.99  & 35.11  & 24.34 \\
\cline{2-9}          & \multirow{6}[2]{*}{GTSRB} & Barlow & 97.67  & 97.02  & 93.92  & 91.14  & 87.06  & 83.85 \\
          &       & BYOL  & 83.33  & 87.52  & 85.68  & 82.21  & 77.18  & 74.78 \\
          &       & DINO  & 87.33  & 90.45  & 88.49  & 86.14  & 83.17  & 80.59 \\
          &       & MoCo2+  & 82.00  & 85.24  & 84.37  & 82.06  & 78.10  & 75.37 \\
          &       & NNCLR & 77.67  & 85.13  & 83.83  & 81.75  & 78.23  & 75.37 \\
          &       & SimCLR & 90.67  & 91.81  & 88.53  & 84.78  & 80.20  & 77.26 \\
    \bottomrule[1.5pt]
    \end{tabular}%
    }
  \label{tab:clean_retrieval}%
     \vspace{-4mm}
\end{table}%

\begin{table*}[htbp]
\setlength{\abovecaptionskip}{2pt}
  \centering
  \caption{The clean accuracy (\%) of different downstream models based on pre-trained encoders over four datasets. PD denotes the pre-training dataset and DD represents the downstream dataset.}
  \scalebox{0.6855}{
    \begin{tabular}{c|c|cccccccccccccc}
    \toprule[1.5pt]
    PD & DD & Barlow & BYOL  & DeepC2 & DINO  & MoCo2+ & MoCo3 & NNCLR & ReSSL & SimCLR & SupCon & SwAV  & VIbCReg & VICReg & W-MSE \\
    \hline
    \multirow{4}[2]{*}{CIFAR10} & CIFAR10 & 93.61  & 94.47  & 90.59  & 91.60  & 95.10  & 94.67  & 93.91  & 92.67  & 93.63  & 96.23  & 92.32  & 92.35  & 93.17  & 90.13  \\
        & STL10 & 83.98  & 83.18  & 77.99  & 83.30  & 84.61  & 83.92  & 83.26  & 81.65  & 81.64  & 84.47  & 81.32  & 81.97  & 82.18  & 78.26  \\
          & GTSRB & 97.57  & 93.66  & 95.99  & 97.04  & 91.90  & 92.09  & 96.28  & 95.97  & 96.59  & 97.39  & 97.77  & 96.96  & 97.11  & 88.63  \\
          & ImageNet & 67.58  & 52.94  & 45.44  & 60.16  & 51.15  & 48.92  & 57.00  & 51.01  & 54.23  & 50.67  & 58.82  & 61.00  & 61.83  & 41.00  \\
    \hline
    \multirow{4}[2]{*}{ImageNet} & CIFAR10 & 72.31  & 72.44  & 69.06  & 73.14  & 73.12  & 72.86  & 73.89  & 72.22  & 69.51  & 70.71  & 71.48  & 72.90  & 73.21  & 68.64  \\
          & STL10 & 65.09  & 64.36  & 61.77  & 65.16  & 65.23  & 63.10  & 65.95  & 65.04  & 62.49  & 63.31  & 64.05  & 65.23  & 64.70  & 61.17  \\
          & GTSRB & 95.49  & 92.18  & 95.16  & 92.28  & 90.28  & 87.88  & 93.93  & 90.82  & 93.71  & 95.81  & 96.18  & 95.13  & 94.51  & 92.66  \\
          & ImageNet & 56.87  & 46.10  & 46.04  & 49.21  & 42.84  & 43.26  & 49.15  & 45.19  & 48.79  & 52.44  & 48.83  & 55.50  & 52.29  & 38.13  \\
    \bottomrule[1.5pt]
    \end{tabular}%
    }
  \label{tab:clean}%
\end{table*}%

\begin{algorithm}[H]
    \caption{Frequency-based Generative Attack Framework}
    \label{optimization_advencoder}

    \begin{algorithmic}[1] 
        \REQUIRE Attacker’s surrogate data point $ x \in \mathcal{D}_{a}$, a pre-trained encoder $g_{\theta}$, max-perturbation constraint $\epsilon$, a fixed noise $z$, adversarial generator parameters  $\theta_{\mathcal{G}}$, hyper-parameters $\alpha$, $\beta$, $\lambda$, temperature parameter $\tau$.
        \ENSURE a universal adversarial noise $\delta$. 
        \STATE {Initiate learning rate $\eta$, batch size $n$.}
        \STATE {Sample a vector $z$ from $\mathcal{N}(0, 1)^{100}$}
        \WHILE {max iterations or not  converge}
            \STATE {$\delta \longleftarrow \mathcal{G}(z)$}
            \STATE {Clip $\delta$ to satisfy imperceptibility constraint $\epsilon$}
            \IF {Adv-PER}
            \STATE {$x^{adv} \longleftarrow x + \delta$}
            \ENDIF
             \IF {Adv-PAT}
            \STATE {$x^{adv} \longleftarrow  x  \odot (1- m) + \delta  \odot m$}
            \ENDIF
            \STATE {Calculate loss mentioned in \cref{eq:3}} 
            \STATE {Update $\mathcal{G}$ through backprop}
        \ENDWHILE
        
    \end{algorithmic} 
\end{algorithm}

\vspace{-4mm}

\section{Supplemental Attack Performance}\label{attack_performance}
In this section, we further investigate the attack performance of AdvEncoder in two types of downstream tasks, classification and retrieval.
We keep all the following experimental settings consistent with the main body.

\subsection{Attack Performance on Classification} \label{attack_performance_classification}
We first provide the normal accuracy of downstream models based on fourteen SSL pre-trained encodes on four different datasets. 
The results in \cref{tab:clean} prove that users using pre-trained encoders can achieve excellent performance on different datasets just by fine-tuning the linear layer.
Specifically, we evaluate AdvEncoder on fourteen victim pre-trained encoders over four downstream tasks using two attacker's surrogate datasets, \textbf{STL10} and \textbf{GTSRB}, respectively.
The performance of our attack using two additional attacker's surrogate datasets on the classification task further confirms that downstream tasks based on pre-trained encoders are exposed to significant security risks. The experimental results  in \cref{tab:attack_performance_per1} and \cref{tab:attack_performance_pat1} demonstrate that an attacker can achieve successful attacks even without prior knowledge of the pre-training dataset and the downstream dataset. 
These findings are in line with the results presented in the main body of the paper.

\begin{table*}[htbp]
\setlength{\abovecaptionskip}{2pt}
  \centering
  \caption{The attack success rate (\%) of Adv-PER under different settings.
  $\mathcal{S}_{1}$ - $\mathcal{S}_{4}$ denote the settings where the downstream datasets are CIFAR10, STL10, GTSRB, ImageNet, respectively, and all the attacker’s surrogate dataset is \textbf{STL10}. $\mathcal{S}_{5}$ -  $\mathcal{S}_{8}$ use \textbf{GTSRB} as the attacker’s surrogate dataset, with the downstream datasets remained the same as $\mathcal{S}_{1}$ - $\mathcal{S}_{4}$. Barlow Twins and DeepCluster v2 are abbreviated as Barlow and DeepC2, respectively.}
   \scalebox{0.7}{
    \begin{tabular}{c|c|cccccccccccccc}
    \toprule[1.5pt]
    Dataset & Setting & Barlow & BYOL  & DeepC2 & DINO  & MoCo2+ & MoCo3 & NNCLR & ReSSL & SimCLR & SupCon & SwAV  & VIbCReg & VICReg & W-MSE \\
    \hline
    \multirow{9}[2]{*}{CIFAR10} & $\mathcal{S}_{1}$    & 89.34  & 88.36  & 86.17  & 89.85  & 78.39  & 87.49  & 90.95  & 88.79  & 70.10 & 90.21  & 51.27  & 89.02  & 81.11  & 62.29  \\
          & $\mathcal{S}_{2}$    & 53.58  & 72.57  & 71.96  & 70.74  & 37.07  & 60.83  & 72.16  & 63.67  & 29.11  & 83.02  & \textcolor[RGB]{169,169,169}{28.52}  & 72.33  & 44.42  & 35.68  \\
          & $\mathcal{S}_{3}$    & \textbf{91.92}  & \textbf{92.05}  & \textbf{89.50}  & 92.96  & \textbf{83.93}  & 82.08  & 91.28  & \textbf{94.46}  & \textbf{70.72}  & 91.99  & 66.02  & 90.76  & \textbf{82.97}  & \textbf{72.55}  \\
          & $\mathcal{S}_{4}$    & 88.09  & 88.50  & 85.74  & 86.43  & 82.08  & \textbf{88.25}  & 87.55  & 85.76  & 68.86  & \textbf{94.48}  & 61.45  & 90.31  & 75.44  & 72.22  \\
          & $\mathcal{S}_{5}$    & 87.81  & 88.85  & 82.75  & 89.71  & 51.37  & 55.67  & 89.22  & 86.83  & 45.38  & 83.38  & 61.40  & 85.78  & 73.91  & 52.28  \\
          & $\mathcal{S}_{6}$    & \textcolor[RGB]{169,169,169}{50.83}  & \textcolor[RGB]{169,169,169}{67.82}  & \textcolor[RGB]{169,169,169}{52.01}  & \textcolor[RGB]{169,169,169}{64.50}  & \textcolor[RGB]{169,169,169}{30.79}  & \textcolor[RGB]{169,169,169}{25.16}  & \textcolor[RGB]{169,169,169}{60.46}  & \textcolor[RGB]{169,169,169}{52.69}  & \textcolor[RGB]{169,169,169}{26.54}  & \textcolor[RGB]{169,169,169}{48.44}  & 33.14  & \textcolor[RGB]{169,169,169}{57.67}  & \textcolor[RGB]{169,169,169}{35.60}  & \textcolor[RGB]{169,169,169}{29.48}  \\
          & $\mathcal{S}_{7}$    & 91.29  & 91.19  & 89.05  & \textbf{94.73}  & 78.57  & 61.55  & \textbf{91.86}  & 94.21  & 64.48  & 80.67  & \textbf{76.86}  & \textbf{91.32}  & 77.77  & 71.33  \\
          & $\mathcal{S}_{8}$    & 82.55  & 90.27  & 79.19  & 83.55  & 66.19  & 65.62  & 83.81  & 79.47  & 61.73  & 79.30  & 66.14  & 80.56  & 64.43  & 63.74  \\
          & AVG   & 79.43  & 84.95  & 79.55  & 84.06  & 63.55  & 65.83  & 83.41  & 80.74  & 54.62  & 81.44  & 55.60  & 82.22  & 66.96  & 57.44  \\
    \hline
    \multirow{9}[2]{*}{ImageNet} & $\mathcal{S}_{1}$    & 61.52  & 77.28  & 62.96  & 67.81  & 68.41  & 61.67  & \textbf{74.31}  & 77.48  & 69.80  & 67.26  & 69.17  & 68.42  & 64.65  & 79.24  \\
          & $\mathcal{S}_{2}$    & 58.01  & 54.82  & \textcolor[RGB]{169,169,169}{46.96}  & 49.93  & 52.23  & \textcolor[RGB]{169,169,169}{52.81}  & 52.77  & 60.24  & 57.91  & 51.44  & 52.33  & 48.33  & 53.47  & 67.53  \\
          & $\mathcal{S}_{3}$    & 62.95  & 72.76  & 63.47  & 71.51  & 71.24  & 68.84  & 65.11  & \textbf{80.42}  & 62.89  & \textbf{71.20}  & 61.69  & 59.40  & 68.19  & 76.03  \\
          & $\mathcal{S}_{4}$    & 69.63  & 71.88  & 69.26  & 66.18  & 67.59  & 64.52  & 67.89  & 73.20  & 72.20  & 68.19  & 71.24  & 64.83  & 69.47  & \textbf{80.28}  \\
          & $\mathcal{S}_{5}$    & \textbf{79.51}  & \textbf{83.14}  & 65.30  & 61.31  & 64.59  & 68.47  & 65.01  & 69.87  & 70.72  & 62.82  & 65.01  & 67.89  & 65.25  & 78.97  \\
          & $\mathcal{S}_{6}$    & \textcolor[RGB]{169,169,169}{55.21}  & \textcolor[RGB]{169,169,169}{50.13}  & 48.68  & \textcolor[RGB]{169,169,169}{46.01}  & \textcolor[RGB]{169,169,169}{49.08}  & 53.54  & \textcolor[RGB]{169,169,169}{49.67}  & \textcolor[RGB]{169,169,169}{49.94}  & \textcolor[RGB]{169,169,169}{51.10}  & \textcolor[RGB]{169,169,169}{44.63}  & \textcolor[RGB]{169,169,169}{49.34}  & \textcolor[RGB]{169,169,169}{46.33}  & \textcolor[RGB]{169,169,169}{48.85}  & \textcolor[RGB]{169,169,169}{58.65}  \\
          & $\mathcal{S}_{7}$    & 79.38  & 77.71  & \textbf{72.36}  & \textbf{77.26}  & \textbf{73.52}  & \textbf{76.74}  & 71.66  & 77.47  & \textbf{75.46}  & 69.55  & \textbf{71.38}  & \textbf{74.10}  & \textbf{70.98}  & 70.72  \\
          & $\mathcal{S}_{8}$    & 72.18  & 68.43  & 68.18  & 65.24  & 62.74  & 64.47  & 64.66  & 65.66  & 68.02  & 64.11  & 68.57  & 67.59  & 66.77  & 73.32  \\
          & AVG   & 67.30  & 69.52  & 62.15  & 63.16  & 63.68  & 63.88  & 63.88  & 69.28  & 66.01  & 62.40  & 63.59  & 62.11  & 63.45  & 73.10  \\
    \bottomrule[1.5pt]
    \end{tabular}%
    }
  \label{tab:attack_performance_per1}%
\end{table*}%

\begin{table*}[htbp]
\setlength{\abovecaptionskip}{2pt}
  \centering
  \caption{The attack success rate (\%) of Adv-PAT under different settings. $\mathcal{S}_{1}$ - $\mathcal{S}_{8}$ represent the same settings as mentioned in \cref{tab:attack_performance_per1}.}
   \scalebox{0.7}{
    \begin{tabular}{c|c|cccccccccccccc}
    \toprule[1.5pt]
    Dataset & Setting & Barlow & BYOL  & DeepC2 & DINO  & MoCo2+ & MoCo3 & NNCLR & ReSSL & SimCLR & SupCon & SwAV  & VIbCReg & VICReg & W-MSE \\
    \hline
    \multirow{9}[2]{*}{CIFAR10} & $\mathcal{S}_{1}$    & \textcolor[RGB]{169,169,169}{83.47}  & 79.55  & 90.88  & 88.61  & \textcolor[RGB]{169,169,169}{81.60}  & 88.83  & \textcolor[RGB]{169,169,169}{65.55}  & 73.08  & 89.91  & 80.03  & 89.30  & 57.27  & 87.58  & 88.61  \\
          & $\mathcal{S}_{2}$    & 88.78  & 80.98  & \textcolor[RGB]{169,169,169}{87.66}  & 79.87  & 82.51  & \textcolor[RGB]{169,169,169}{77.50}  & 75.83  & 73.19  & 89.32  & 69.42  & \textcolor[RGB]{169,169,169}{81.41}  & 56.82  & 82.36  & 81.32  \\
          & $\mathcal{S}_{3}$    & 93.14  & 89.95  & 95.44  & 86.40  & \textbf{99.08}  & 92.43  & 89.84  & 88.62  & 94.98  & 88.27  & 97.09  & 86.43  & 94.47  & 88.97  \\
          & $\mathcal{S}_{4}$    & 93.41  & \textbf{98.03}  & 99.53  & \textbf{98.16}  & 98.55  & 97.15  & 94.25  & \textbf{97.85}  & \textbf{98.97}  & \textbf{96.29}  & \textbf{98.53}  & \textbf{94.82}  & \textbf{97.76}  & \textbf{96.33}  \\
          & $\mathcal{S}_{5}$    & 83.58  & 87.64  & 90.88  & 82.59  & 86.05  & 89.83  & 67.80  & \textcolor[RGB]{169,169,169}{64.38}  & 89.90  & 77.01  & 89.30  & 53.89  & \textcolor[RGB]{169,169,169}{75.92}  & 87.02  \\
          & $\mathcal{S}_{6}$    & 83.79  & \textcolor[RGB]{169,169,169}{78.71}  & 89.67  & \textcolor[RGB]{169,169,169}{71.72}  & 82.41  & 83.64  & 66.58  & 75.40  & \textcolor[RGB]{169,169,169}{88.42}  & \textcolor[RGB]{169,169,169}{56.32}  & 83.92  & \textcolor[RGB]{169,169,169}{52.26}  & {78.69}  & \textcolor[RGB]{169,169,169}{79.55}  \\
          & $\mathcal{S}_{7}$    & 92.84  & 88.83  & 94.53  & 87.63  & 98.27  & 92.41  & 87.88  & 88.94  & 95.23  & 84.60  & 97.09  & 84.20  & 90.52  & 86.22  \\
          & $\mathcal{S}_{8}$    & \textbf{93.64}  & 96.68  & \textbf{99.70}  & 95.97  & 98.25  & \textbf{97.32}  & \textbf{94.45}  & 97.80  & 98.94  & 94.43  & 98.17  & 94.10  & 94.59  & 95.01  \\
          & AVG   & 89.08  & 87.55  & 93.53  & 86.37  & 90.84  & 89.89  & 80.27  & 82.41  & 93.21  & 80.80  & 91.85  & 72.47  & 87.74  & 87.88  \\
    \hline
    \multirow{9}[2]{*}{ImageNet} & $\mathcal{S}_{1}$    & 89.33  & 88.45  & \textcolor[RGB]{169,169,169}{89.22}  & 89.41  & \textcolor[RGB]{169,169,169}{87.28}  & \textcolor[RGB]{169,169,169}{88.80}  & 88.76  & 92.01  & 90.31  & 90.50  & \textcolor[RGB]{169,169,169}{90.06}  & 89.04  & 89.18  & 91.15  \\
          & $\mathcal{S}_{2}$    & \textcolor[RGB]{169,169,169}{83.53}  & 88.11  & 89.98  & \textcolor[RGB]{169,169,169}{89.09}  & 90.65  & 91.22  & 88.86  & 91.11  & \textcolor[RGB]{169,169,169}{89.26}  & 90.92  & 90.28  & \textcolor[RGB]{169,169,169}{86.13}  & 87.55  & \textcolor[RGB]{169,169,169}{89.84}  \\
          & $\mathcal{S}_{3}$    & 93.37  & \textbf{99.19}  & 97.04  & 94.95  & 95.54  & \textbf{98.47}  & 98.36  & 90.67  & 94.33  & 94.52  & 97.07  & 95.42  & 95.31  & 98.30  \\
          & $\mathcal{S}_{4}$    & \textbf{98.60}  & 98.63  & 99.21  & \textbf{98.79}  & 98.06  & 98.30  & \textbf{98.40}  & 98.17  & \textbf{99.02}  & \textbf{99.15}  & 98.66  & \textbf{98.59}  & \textbf{98.79}  & \textbf{98.49}  \\
          & $\mathcal{S}_{5}$    & 86.45  & 88.95  & \textcolor[RGB]{169,169,169}{89.22}  & 89.41  & 91.26  & 89.04  & 88.72  & 92.00  & 90.30  & 90.50  & \textcolor[RGB]{169,169,169}{90.06}  & 89.03  & 89.49  & 91.19  \\
          & $\mathcal{S}_{6}$    & 87.00  & \textcolor[RGB]{169,169,169}{87.15}  & 89.98  & 89.48  & 90.58  & 91.50  & \textcolor[RGB]{169,169,169}{88.15}  & 91.19  & 89.60  & \textcolor[RGB]{169,169,169}{90.21}  & 90.15  & 89.88  & \textcolor[RGB]{169,169,169}{87.54}  & 89.86  \\
          & $\mathcal{S}_{7}$    & 94.05  & 97.37  & 96.72  & 94.65  & 96.54  & 97.43  & 98.36  & \textcolor[RGB]{169,169,169}{90.65}  & 94.30  & 92.31  & 97.09  & 96.29  & 96.02  & 98.33  \\
          & $\mathcal{S}_{8}$    & 98.54  & 98.61  & \textbf{99.22}  & 98.43  & \textbf{98.30}  & 98.19  & 98.25  & \textbf{98.32}  & \textbf{99.02}  & 98.38  & \textbf{98.88}  & 98.56  & 98.77  & 98.43  \\
          & AVG   & 91.36  & 93.31  & 93.82  & 93.03  & 93.52  & 94.12  & 93.48  & 93.01  & 93.27  & 93.31  & 94.03  & 92.87  & 92.83  & 94.45  \\
    \bottomrule[1.5pt]
    \end{tabular}%
    }
  \label{tab:attack_performance_pat1}%
\end{table*}%

\begin{table*}[htbp]
\setlength{\abovecaptionskip}{2pt}
  \centering
   \caption{The retrieval attack performance (\%) of AdvEncoder under different settings on the pre-training dataset CIFAR10. PD denotes the pre-training dataset, SD indicates the attacker’s surrogate dataset  and DD represents the downstream dataset. }
        \scalebox{0.675}{
    \begin{tabular}{c|c|c|c|cccccccccccc}
    \toprule[1.5pt]
    \multirow{2}{*}{PD} & \multirow{2}{*}{SD} & \multirow{2}{*}{DD} & \multirow{2}{*}{Model} & \multicolumn{2}{c}{top1} & \multicolumn{2}{c}{top5} & \multicolumn{2}{c}{top10} & \multicolumn{2}{c}{top20} & \multicolumn{2}{c}{top50} & \multicolumn{2}{c}{top100} \\
          &       &       &       & p\_mAP & pat\_mAP & p\_mAP & pat\_mAP & p\_mAP & pat\_mAP & p\_mAP & pat\_mAP & p\_mAP & pat\_mAP & p\_mAP & pat\_mAP \\
     \hline
    \multirow{24}[8]{*}{CIFAR10} & \multirow{12}[4]{*}{CIFAR10} & \multirow{6}[2]{*}{STL10} & Barlow & 8.33  & 11.00  & 21.03  & 23.80  & 22.90  & 24.97  & 21.75  & 21.54  & 16.53  & 16.89  & 14.13  & 14.10  \\
          &       &       & BYOL  & 10.00  & 10.00  & 19.89  & 18.15  & 21.96  & 20.65  & 20.79  & 19.82  & 16.83  & 16.44  & 14.26  & 13.96  \\
          &       &       & DINO  & 10.00  & 9.00  & 19.13  & 19.26  & 20.85  & 21.39  & 20.07  & 20.22  & 16.50  & 16.45  & 13.61  & 14.27  \\
          &       &       & MoCo2+  & 12.00  & 9.33  & 21.88  & 21.00  & 22.74  & 23.28  & 20.96  & 23.05  & 16.61  & 17.08  & 14.22  & 14.25  \\
          &       &       & NNCLR & 9.00  & 9.33  & 18.29  & 20.31  & 20.49  & 21.85  & 19.46  & 21.51  & 16.13  & 16.99  & 13.85  & 14.13  \\
          &       &       & SimCLR & 10.67  & 9.67  & 21.55  & 17.14  & 23.39  & 20.40  & 21.16  & 19.69  & 17.05  & 16.22  & 14.22  & 14.13  \\
\cline{3-16}          &       & \multirow{6}[2]{*}{GTSRB} & Barlow & 18.33  & 10.00  & 21.57  & 10.00  & 22.21  & 10.00  & 21.83  & 10.05  & 21.30  & 10.06  & 20.92  & 10.07  \\
          &       &       & BYOL  & 14.33  & 14.33  & 15.64  & 17.41  & 15.56  & 17.43  & 15.61  & 17.66  & 15.17  & 17.91  & 14.83  & 17.77  \\
          &       &       & DINO  & 12.67  & 17.33  & 18.13  & 25.29  & 18.22  & 24.75  & 17.82  & 24.06  & 17.13  & 23.57  & 16.17  & 23.36  \\
          &       &       & MoCo2+  & 30.00  & 12.33  & 34.01  & 12.34  & 33.86  & 12.24  & 32.81  & 12.21  & 30.93  & 12.17  & 29.37  & 12.08  \\
          &       &       & NNCLR & 16.33  & 18.67  & 18.16  & 22.88  & 17.64  & 23.18  & 17.28  & 22.86  & 16.94  & 22.77  & 16.87  & 22.79  \\
          &       &       & SimCLR & 46.00  & 10.00  & 50.10  & 10.33  & 49.21  & 10.38  & 48.37  & 10.49  & 46.68  & 10.43  & 45.26  & 10.41  \\
\cline{2-16}          & \multirow{12}[4]{*}{ImageNet} & \multirow{6}[2]{*}{STL10} & Barlow & 11.00  & 11.00  & 20.03  & 21.01  & 23.10  & 22.43  & 21.75  & 20.77  & 17.43  & 16.64  & 14.45  & 14.20  \\
          &       &       & BYOL  & 12.67  & 10.00  & 23.25  & 17.04  & 24.60  & 18.83  & 22.69  & 20.95  & 17.37  & 17.14  & 14.55  & 14.25  \\
          &       &       & DINO  & 8.00  & 10.00  & 18.41  & 18.76  & 20.01  & 20.51  & 19.00  & 20.58  & 15.78  & 16.34  & 13.43  & 14.00  \\
          &       &       & MoCo2+  & 10.67  & 8.33  & 20.19  & 16.30  & 22.34  & 17.62  & 20.48  & 18.27  & 16.85  & 15.63  & 14.29  & 13.62  \\
          &       &       & NNCLR & 8.33  & 11.33  & 18.97  & 22.09  & 22.26  & 24.21  & 20.94  & 22.41  & 16.14  & 17.40  & 13.83  & 14.19  \\
          &       &       & SimCLR & 9.33  & 11.33  & 19.62  & 16.52  & 21.24  & 19.95  & 20.30  & 18.47  & 16.10  & 16.27  & 13.82  & 14.24  \\
\cline{3-16}          &       & \multirow{6}[2]{*}{GTSRB} & Barlow & 12.67  & 16.00  & 17.05  & 17.79  & 17.08  & 17.76  & 15.48  & 16.79  & 14.21  & 16.37  & 13.63  & 16.08  \\
          &       &       & BYOL  & 13.00  & 10.33  & 17.36  & 12.83  & 16.71  & 13.37  & 16.31  & 13.33  & 15.93  & 13.58  & 15.39  & 14.03  \\
          &       &       & DINO  & 14.00  & 26.00  & 19.37  & 32.69  & 19.17  & 33.77  & 18.78  & 33.33  & 18.57  & 32.10  & 18.39  & 31.49  \\
          &       &       & MoCo2+  & 27.67  & 15.00  & 31.15  & 15.04  & 31.58  & 14.81  & 31.37  & 14.76  & 30.86  & 14.54  & 30.42  & 14.24  \\
          &       &       & NNCLR & 8.67  & 20.33  & 10.18  & 24.39  & 10.32  & 24.19  & 10.55  & 23.32  & 11.17  & 22.67  & 11.22  & 22.30  \\
          &       &       & SimCLR & 13.00  & 10.00  & 21.82  & 10.00  & 22.07  & 10.04  & 22.14  & 10.04  & 22.05  & 10.04  & 21.58  & 10.11  \\
    \bottomrule[1.5pt]
    \end{tabular}%
    }
  \label{tab:retrieval_attack_cifar10}%
\end{table*}%
 
\begin{table*}[htbp]
\setlength{\abovecaptionskip}{2pt}
  \centering
   \caption{The retrieval attack performance (\%) of AdvEncoder under different settings on the pre-training dataset ImageNet}
     \scalebox{0.675}{
    \begin{tabular}{c|c|c|c|cccccccccccc}
    \toprule[1.5pt]
    \multirow{2}{*}{PD} & \multirow{2}{*}{SD} & \multirow{2}{*}{DD} & \multirow{2}{*}{Model} & \multicolumn{2}{c}{top1} & \multicolumn{2}{c}{top5} & \multicolumn{2}{c}{top10} & \multicolumn{2}{c}{top20} & \multicolumn{2}{c}{top50} & \multicolumn{2}{c}{top100} \\
          &       &       &       & p\_mAP & pat\_mAP & p\_mAP & pat\_mAP & p\_mAP & pat\_mAP & p\_mAP & pat\_mAP & p\_mAP & pat\_mAP & p\_mAP & pat\_mAP \\
    \hline
    \multirow{24}[8]{*}{ImageNet} & \multirow{12}[4]{*}{CIFAR10} & \multirow{6}[2]{*}{STL10} & Barlow & 11.67  & 9.00  & 21.70  & 19.44  & 23.26  & 21.15  & 21.55  & 20.68  & 17.02  & 16.36  & 14.27  & 13.76  \\
          &       &       & BYOL  & 9.67  & 8.00  & 20.25  & 18.01  & 21.76  & 19.64  & 21.03  & 19.26  & 16.36  & 16.84  & 13.97  & 14.03  \\
          &       &       & DINO  & 9.67  & 9.33  & 22.10  & 19.82  & 22.95  & 21.12  & 22.01  & 19.93  & 17.18  & 16.10  & 14.53  & 14.15  \\
          &       &       & MoCo2+  & 11.00  & 14.00  & 20.73  & 23.04  & 22.77  & 26.89  & 21.19  & 25.05  & 16.36  & 20.26  & 13.90  & 14.68  \\
          &       &       & NNCLR & 12.33  & 10.67  & 23.82  & 22.85  & 24.99  & 26.22  & 22.92  & 24.89  & 17.90  & 17.20  & 14.98  & 13.60  \\
          &       &       & SimCLR & 9.67  & 11.33  & 19.80  & 21.07  & 21.76  & 23.59  & 20.13  & 21.46  & 16.62  & 17.04  & 14.31  & 13.99  \\
\cline{3-16}          &       & \multirow{6}[2]{*}{GTSRB} & Barlow & 38.67  & 10.00  & 43.57  & 11.16  & 42.81  & 11.63  & 41.57  & 11.67  & 39.82  & 11.22  & 38.66  & 11.21  \\
          &       &       & BYOL  & 40.67  & 10.00  & 46.60  & 10.07  & 45.81  & 10.07  & 43.88  & 10.07  & 41.21  & 10.24  & 39.33  & 10.15  \\
          &       &       & DINO  & 26.67  & 10.00  & 30.54  & 10.07  & 30.72  & 10.17  & 29.85  & 10.24  & 29.45  & 10.55  & 28.69  & 10.60  \\
          &       &       & MoCo2+  & 32.33  & 7.00  & 39.30  & 11.07  & 38.93  & 11.75  & 37.79  & 12.25  & 35.51  & 11.55  & 34.81  & 11.23  \\
          &       &       & NNCLR & 30.67  & 9.00  & 36.07  & 11.96  & 36.14  & 11.66  & 35.22  & 11.97  & 33.33  & 12.04  & 32.50  & 12.16  \\
          &       &       & SimCLR & 40.33  & 10.00  & 45.80  & 10.42  & 45.56  & 13.63  & 44.11  & 15.75  & 42.12  & 14.76  & 40.88  & 14.68  \\
\cline{2-16}          & \multirow{12}[4]{*}{ImageNet} & \multirow{6}[2]{*}{STL10} & Barlow & 13.00  & 9.67  & 22.27  & 19.99  & 23.02  & 21.75  & 21.45  & 20.18  & 16.92  & 16.28  & 14.09  & 13.76  \\
          &       &       & BYOL  & 11.33  & 11.33  & 20.36  & 20.41  & 22.14  & 23.85  & 21.46  & 20.93  & 16.94  & 15.90  & 14.03  & 13.51  \\
          &       &       & DINO  & 9.00  & 9.33  & 19.75  & 20.21  & 21.46  & 22.75  & 21.04  & 20.28  & 16.44  & 16.29  & 13.90  & 14.45  \\
          &       &       & MoCo2+  & 9.67  & 8.00  & 20.40  & 19.23  & 21.12  & 22.28  & 20.02  & 23.67  & 16.33  & 17.06  & 13.82  & 13.80  \\
          &       &       & NNCLR & 11.00  & 11.33  & 20.59  & 19.96  & 22.82  & 24.91  & 20.74  & 23.10  & 16.89  & 16.33  & 14.17  & 13.62  \\
          &       &       & SimCLR & 9.33  & 9.00  & 19.36  & 21.31  & 21.43  & 23.91  & 20.83  & 20.59  & 16.84  & 16.37  & 14.15  & 13.64  \\
\cline{3-16}          &       & \multirow{6}[2]{*}{GTSRB} & Barlow & 28.67  & 9.33  & 34.08  & 10.68  & 34.65  & 11.10  & 34.38  & 11.61  & 34.14  & 11.99  & 33.92  & 12.08  \\
          &       &       & BYOL  & 35.33  & 10.00  & 41.06  & 10.00  & 41.40  & 10.00  & 38.94  & 10.12  & 36.42  & 10.30  & 34.38  & 10.22  \\
          &       &       & DINO  & 27.00  & 10.00  & 31.89  & 10.12  & 31.57  & 10.05  & 30.65  & 10.09  & 29.95  & 10.11  & 29.18  & 10.06  \\
          &       &       & MoCo2+  & 22.33  & 12.33  & 33.48  & 13.18  & 32.40  & 13.73  & 30.93  & 13.29  & 29.12  & 12.48  & 27.94  & 12.07  \\
          &       &       & NNCLR & 31.67  & 10.00  & 36.82  & 12.05  & 37.07  & 13.04  & 37.21  & 13.03  & 35.99  & 13.29  & 35.20  & 12.54  \\
          &       &       & SimCLR & 19.00  & 10.00  & 21.68  & 11.83  & 21.58  & 12.41  & 21.42  & 12.72  & 20.90  & 12.93  & 20.36  & 13.37  \\
    \bottomrule[1.5pt]
    \end{tabular}%
    }
  \label{tab:retrieval_attack_imagenet}%
\end{table*}%

\subsection{Attack Performance on Retrieval}  \label{attack_performance_retrieval}

 We aim to investigate the impact of AdvEncoder on the retrieval accuracy of downstream tasks under different settings. 
 We select Barlow Twins\cite{zbontar2021barlow}, MoCo v2+\cite{chen2020improved}, BYOL\cite{grill2020bootstrap}, DINO\cite{caron2021emerging}, NNCLR\cite{dwibedi2021little}, and SimCLR \cite{chen2020simple} as pre-trained encoders to evaluate the performance of AdvEncoder in retrieval downstream tasks.
We use CIFAR10 and ImageNet as pre-training datasets for the victim models. As shown in \cref{tab:clean_retrieval}, we provide the precision (mAP) corresponding to their different settings on STL10 and GTSRB downstream retrieval tasks. We then use CIFAR10 and ImageNet as the surrogate datasets for the attacker to launch attacks on the downstream tasks.
We use per-mAP and pat-mAP metrics, where lower values indicate better attack performance. 
The results in ~\cref{tab:retrieval_attack_cifar10} and ~\cref{tab:retrieval_attack_imagenet} illustrate that AdvEncoder can successfully attack the downstream retrieval task without any knowledge of the pre-training and downstream datasets.

\begin{figure*}[!t]   
\setlength{\abovecaptionskip}{2pt}
\centering
  \centering
     \subcaptionbox{Object Detection}{\includegraphics[width=0.23\textwidth]{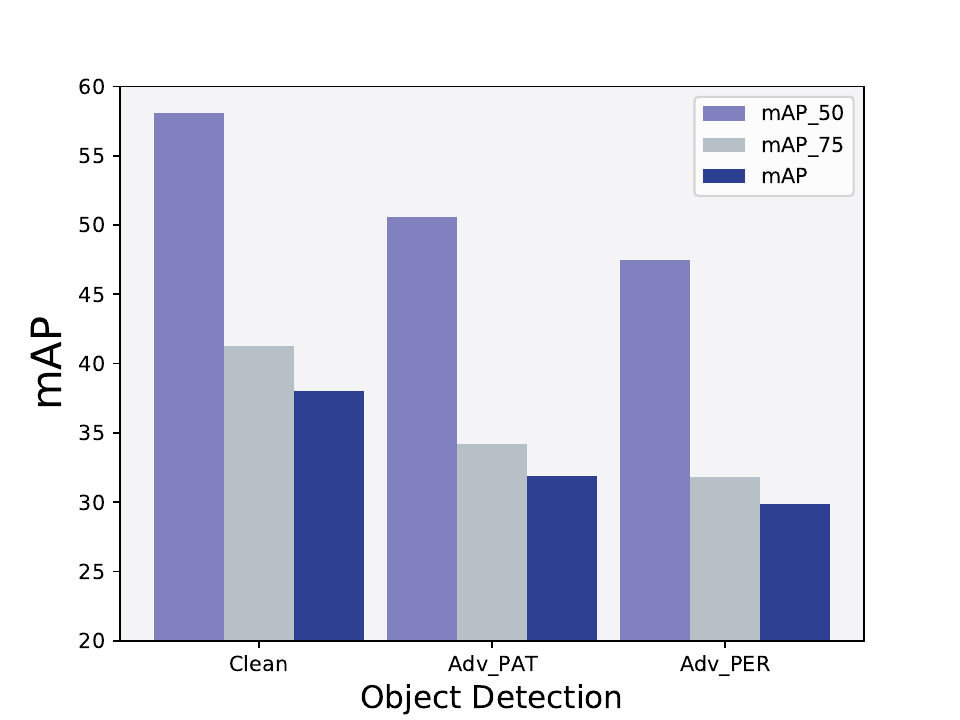}}
      \subcaptionbox{Semantic Segmentation}{\includegraphics[width=0.23\textwidth]{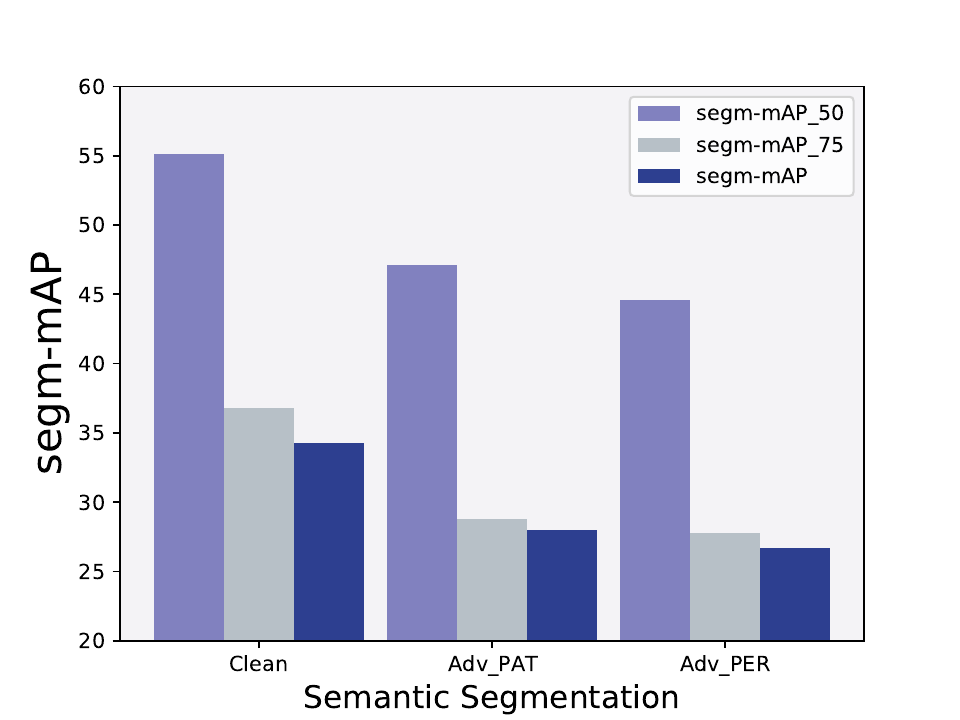}}
      \subcaptionbox{Random Seed}
      {\includegraphics[width=0.235\textwidth]{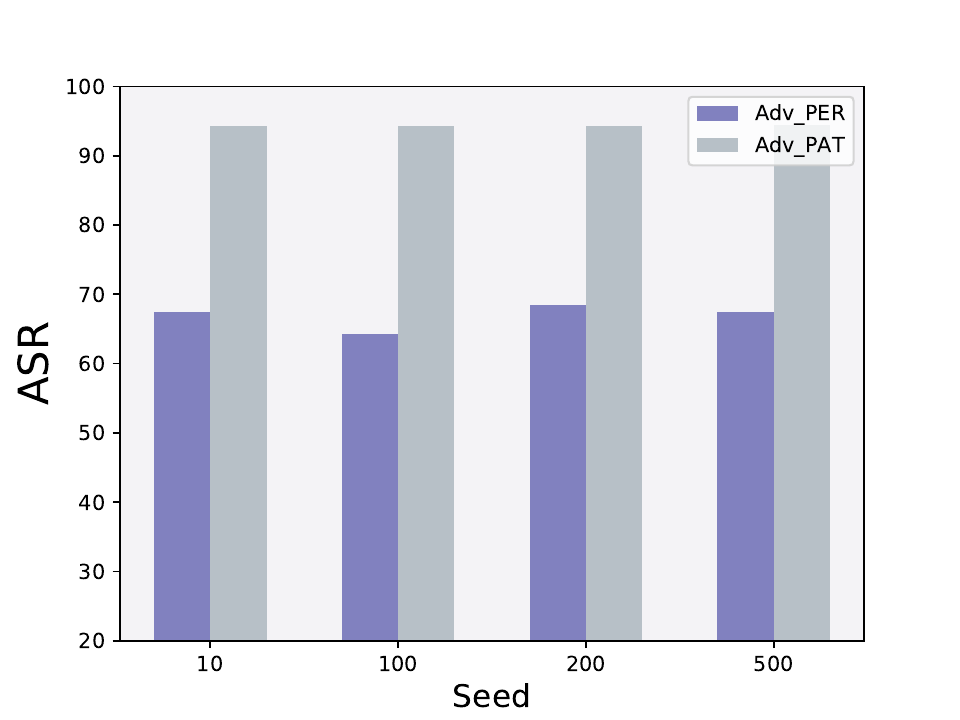}}
       \subcaptionbox{Backbone}{\includegraphics[width=0.235\textwidth]{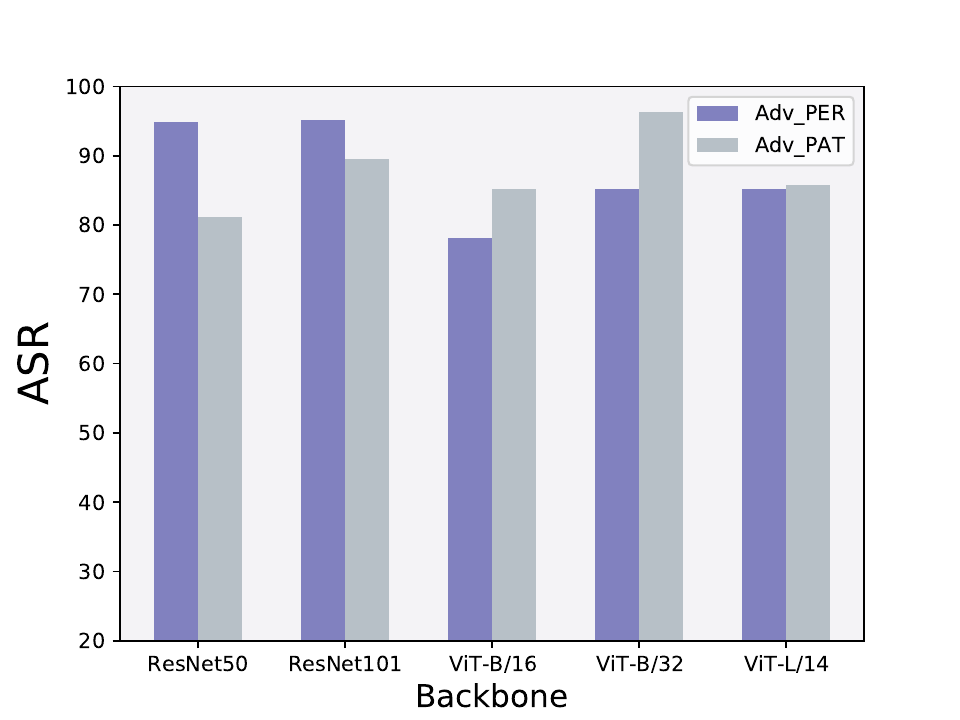}}
      \caption{
      The attack performance (\%) of AdvEncoder in different experimental scenarios}
       \vspace{-0.2cm}
       \label{fig:retriveal1}
\end{figure*}

\begin{figure*}[!t]   
\centering
  \centering
    \subcaptionbox{C2I-GTS-PAT}{\includegraphics[width=0.33\textwidth]{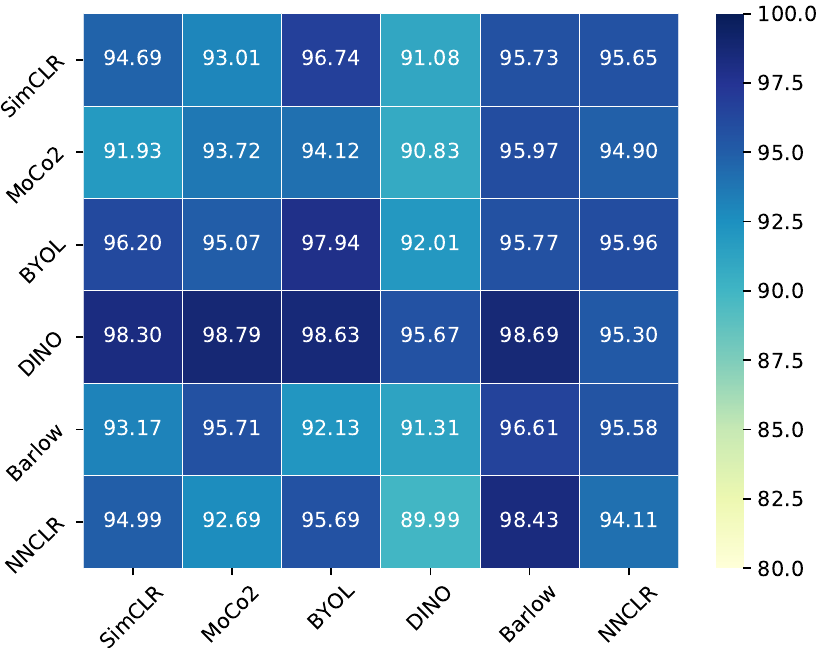}}
      \subcaptionbox{I2I-GTS-PAT}{\includegraphics[width=0.33\textwidth]{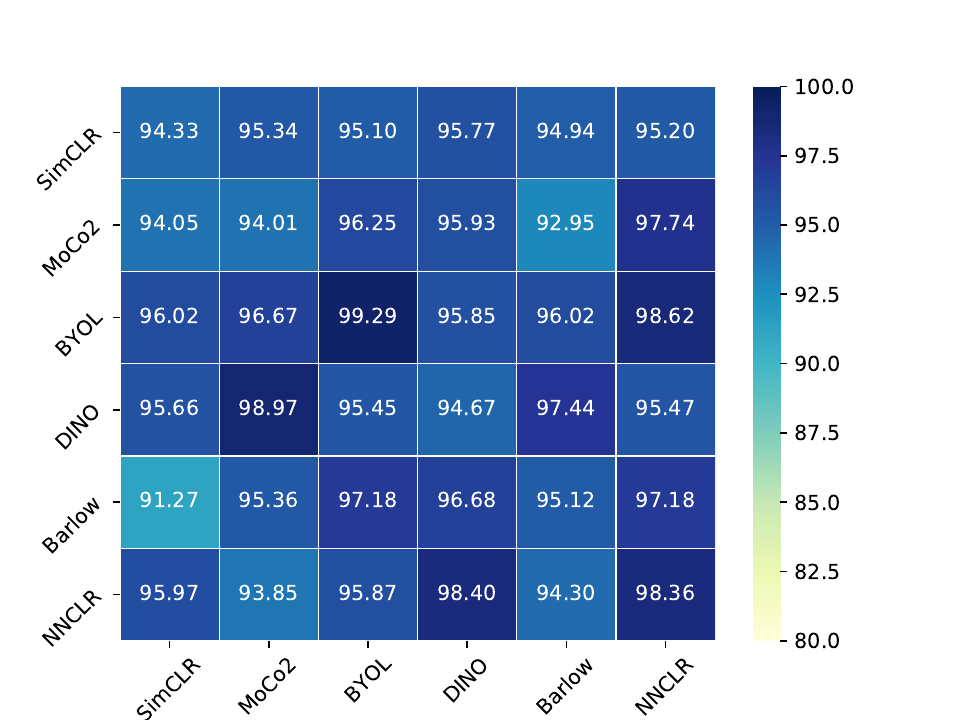}}
    \subcaptionbox{I2I-STL-PER}{\includegraphics[width=0.32\textwidth]{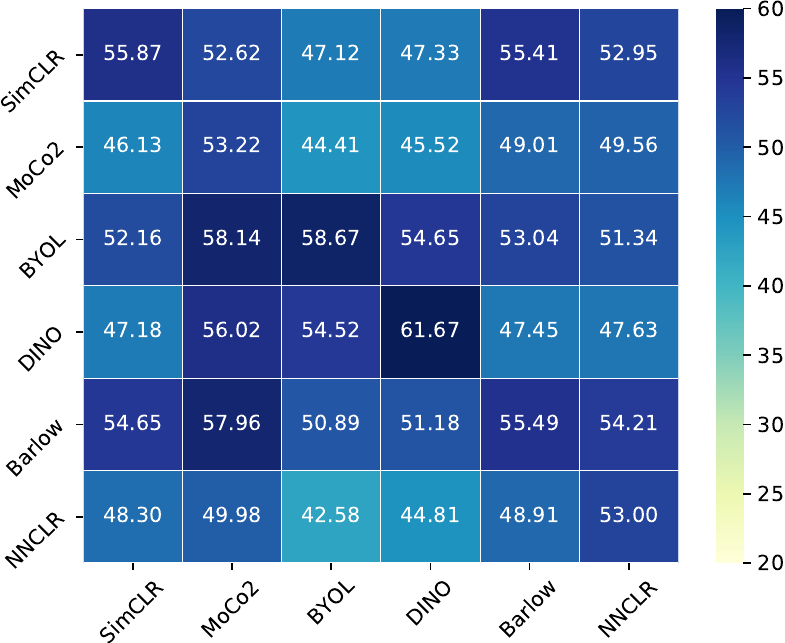}}
      \subcaptionbox{C2I-STL-PER}{\includegraphics[width=0.32\textwidth]{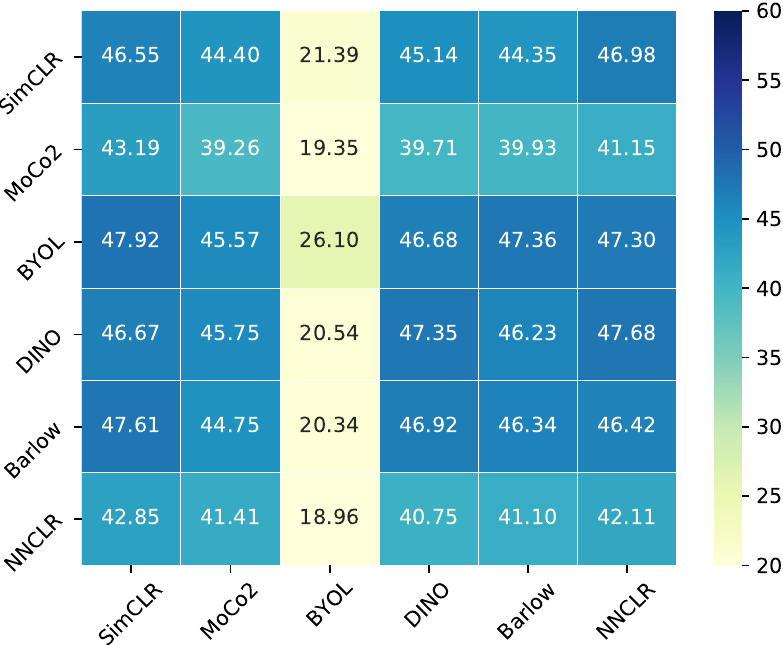}}
    \subcaptionbox{I2I-STL-PAT}{\includegraphics[width=0.335\textwidth]{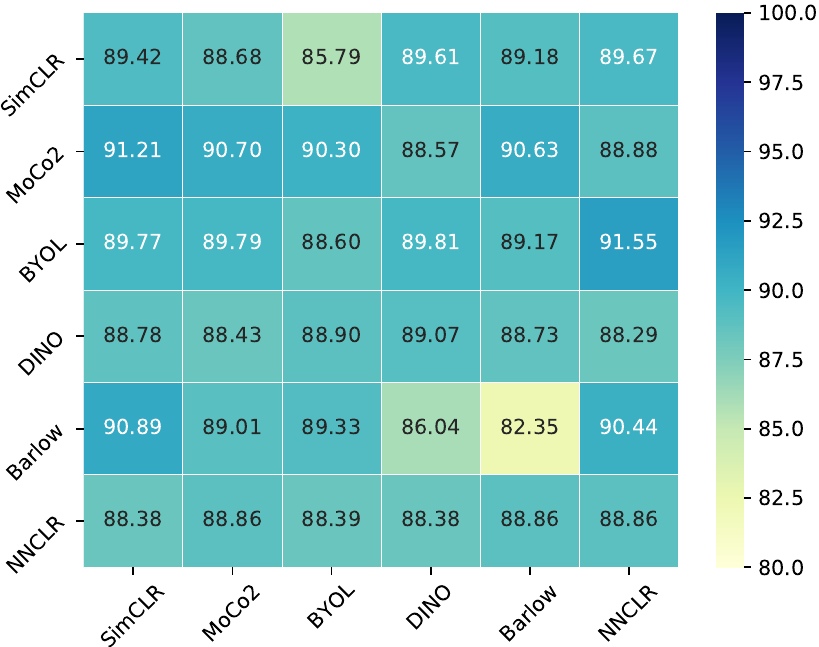}}
    \subcaptionbox{C2I-STL-PAT}{\includegraphics[width=0.33\textwidth]{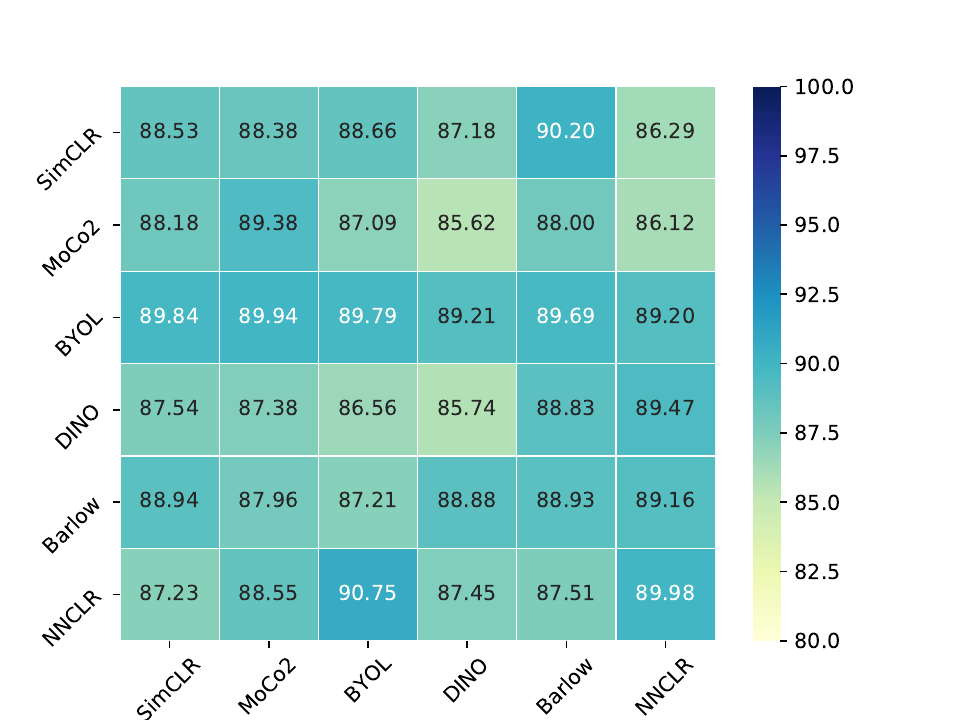}}
      \caption{The attack success rate (\%) of transferability crossing pre-training datasets and  SSL methods
      }
       \label{fig:transfer}
           \vspace{-0.2cm}
\end{figure*}

\subsection{Attack Performance on Object Detection \& Semantic Segmentation}  
We provide the attack performance of AdvEncoder in  \textbf{Object Detection} and \textbf{Semantic Segmentation} tasks using ImageNet as the surrogate dataset in \cref{fig:retriveal1}(a) - (b). 
We employe the official MOCOv2 model based on ResNet50 and fine-tune it on the COCO dataset using Mask R-CNN for the above two types of tasks.
The results in~\cref{fig:retriveal1}(a) - (b) illustrate that AdvEncoder can successfully attack the two types of downstream tasks.

\section{Supplemental Ablation Study} \label{ablation}
In this section, we explore the effect of different random seeds and backbones on the attack performance of AdvEncoder. 
The following experimental settings are consistent with the main body. 
We choose CIFAR10 as the surrogate dataset and GTSRB as the downstream dataset.

\noindent\textbf{The Effect of Random Seed.} 
The default random number seed for our experiments is 100, and we further provide results for different randomized seeds in \cref{fig:retriveal1}(c).

\noindent\textbf{The Effect of Backbone.}
We provide ASRs for downstream GTSRB tasks for five architectures of CLIP (ResNet50, ResNet101, ViT-B/16, ViT-B/32, ViT-L/14).
The results in \cref{fig:retriveal1}(d) show that AdvEncoder can successfully attack downstream tasks based on the pre-trained encoders with different backbones.
 \vspace{-4mm}

\section{Supplemental Transferability Study}  \label{transferability}

In this section, we aim to investigate the transferability of AdvEncoder from two distinct perspectives: pre-training datasets and crossing SSL methods. The experimental settings in this analysis are consistent with those outlined in the main body of the paper. 
From~\cref{fig:transfer}, we explore the attack performance of Adv-PAT and Adv-PER in different transportability scenarios.
C2I-GTS-PAT represents the two encoders we trained using CIFAR10 and ImageNet, on which we made adversarial examples of Adv-PAT and downstream tasks of GTSRB, respectively. I2I-STL-PER represents the two encoders we trained using ImageNet and ImageNet, on which we made adversarial examples of Adv-PER and downstream tasks of STL10, respectively.
The other captions have the same definition.
We can see that Adv-Encoder has good transferability between different downstream tasks based on different encoders.

\section{Visualization}\label{Visualization}
\begin{figure*}[!t]   
\centering
  \centering
  \subcaptionbox{Adv-PER}{\includegraphics[width=0.95\textwidth]{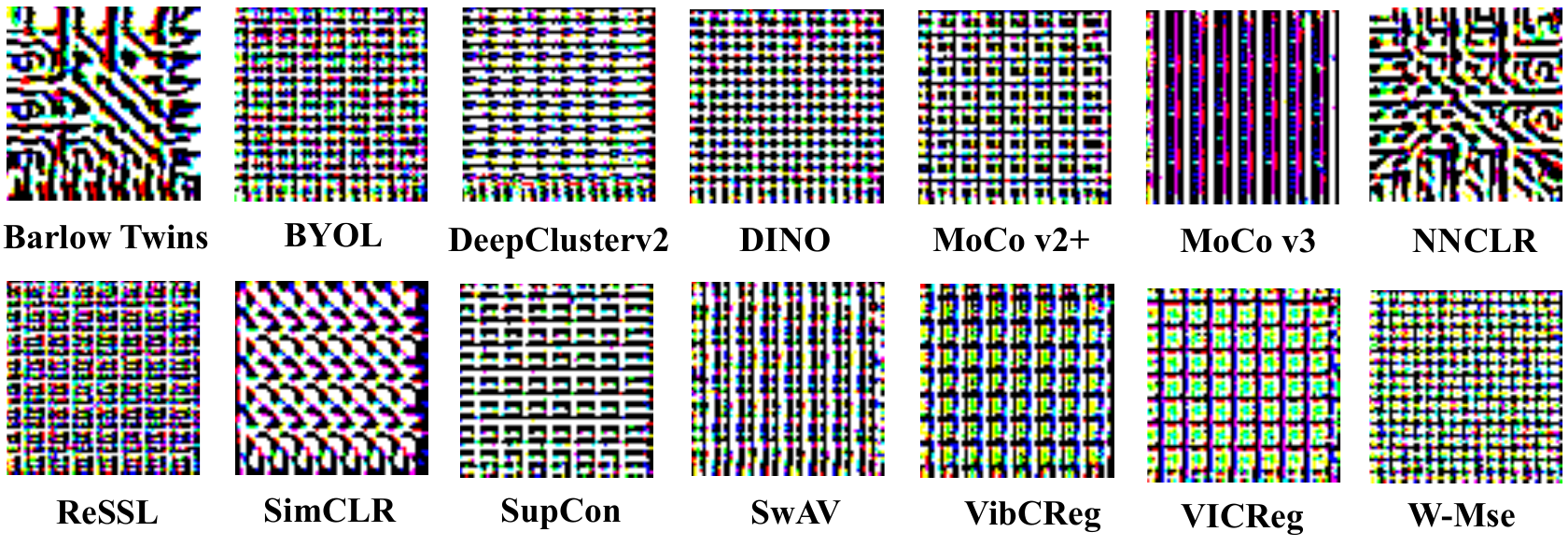}}
    \subcaptionbox{Adv-PAT}{\includegraphics[width=0.95\textwidth]{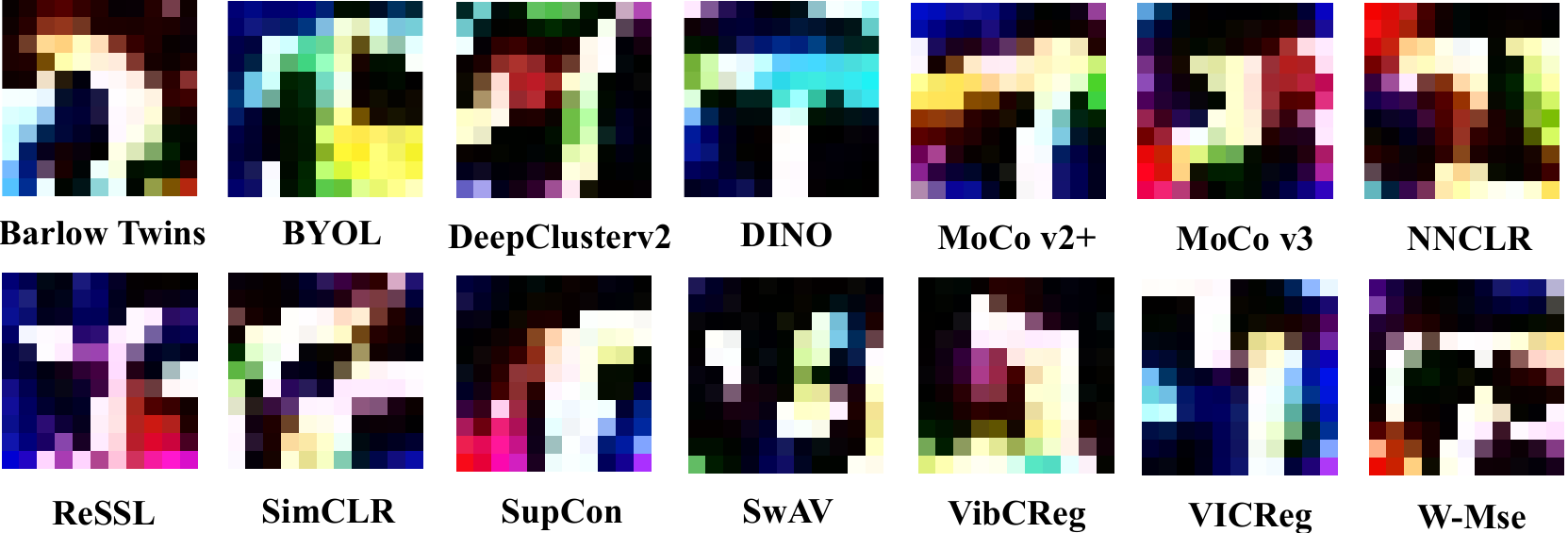}}
    \caption{
    Visualization of universal adversarial perturbations and patches generated by AdvEncoder based on fourteen SSL encoders trained on Imagenet with CIFAR10 as the attacker’s surrogate dataset
      }
       \label{fig:visualization_advencoder}
           \vspace{-0.2cm}
\end{figure*}

In this section, as shown in the \cref{fig:visualization_advencoder}, we show adversarial perturbations and patches generated by AdvEncoder using the attacker’s surrogate dataset CIFAR10 for each of the fourteen SSL encoders trained with ImageNet.

\end{document}